\documentclass[acmtog,nonacm]{acmart}

\setcopyright{none}
\acmDOI{}
\acmISBN{}
\acmConference{}{}{}
\acmVolume{}
\acmNumber{}
\acmArticle{}
\acmYear{}
\acmMonth{}

\renewcommand\footnotetextcopyrightpermission[1]{}

\AtBeginDocument{%
  \fancypagestyle{firstpagestyle}{%
    \fancyhf{}%
    \fancyhead[R]{\thepage}%
  }%
  \fancypagestyle{standardpagestyle}{%
    \fancyhf{}%
    \fancyhead[RO,LE]{\thepage}%
  }%
  \pagestyle{standardpagestyle}%
}

\hypersetup{
  pdftitle={MeshLoom: Feed-Forward Non-Rigid Registration of Mesh Sequences},
  pdfauthor={Jianqi Chen, Jiraphon Yenphraphai, Xiangjun Tang, Sergey Tulyakov, Chaoyang Wang, Peter Wonka, Rameen Abdal},
  pdfsubject={Preprint},
  pdfkeywords={Non-rigid mesh registration, dynamic meshes, dense correspondence, mesh deformation, motion interpolation},
  pdfcreator={LaTeX},
  pdfproducer={}
}

\usepackage{booktabs}
\usepackage[table,xcdraw]{xcolor}
\usepackage{multirow}
\usepackage{enumitem}

\newcommand\awesomemesh{{{MeshLoom}}\xspace}
\newcommand\eg{\textit{e.g.}}
\newcommand\ie{\textit{i.e.}}

\usepackage[ruled]{algorithm2e}

\SetAlFnt{\small}
\SetAlCapFnt{\small}
\SetAlCapNameFnt{\small}
\SetAlCapHSkip{0pt}

\usepackage[normalem]{ulem}
\useunder{\uline}{\ul}{}

\begin{document}
\title{\awesomemesh: Feed-Forward Non-Rigid Registration of Mesh Sequences}

\author{Jianqi Chen}
\email{jianqi.chen@kaust.edu.sa}
\affiliation{%
  \institution{KAUST}
  \country{Saudi Arabia}
}
\affiliation{%
  \institution{Snap Inc.}
  \country{United States of America}
}

\author{Jiraphon Yenphraphai}
\email{jyenphra@purdue.edu}
\affiliation{%
  \institution{Purdue University}
  \country{United States of America}
}
\affiliation{%
  \institution{Snap Inc.}
  \country{United States of America}
}

\author{Xiangjun Tang}
\email{xiangjun.tang@outlook.com}
\affiliation{%
  \institution{KAUST}
  \country{Saudi Arabia}
}

\author{Sergey Tulyakov}
\email{stulyakov@snapchat.com}
\affiliation{%
  \institution{Snap Inc.}
  \country{United States of America}
}

\author{Chaoyang Wang}
\email{gordon.w.1991@gmail.com}
\affiliation{%
  \institution{Snap Inc.}
  \country{United States of America}
}

\author{Peter Wonka}
\email{pwonka@gmail.com}
\affiliation{%
  \institution{KAUST}
  \country{Saudi Arabia}
}
\affiliation{%
  \institution{Snap Inc.}
  \country{United States of America}
}

\author{Rameen Abdal}
\email{rabdal@snapchat.com}
\affiliation{%
  \institution{Snap Inc.}
  \country{United States of America}
}

\begin{abstract}

We present \awesomemesh, a feed-forward registration network that directly reconstructs vertex deformations across mesh sequences. Our approach advances non-rigid registration beyond existing models, which are typically constrained by costly per-instance optimization, narrow object categories, pairwise-only inputs, or merely intermediate outputs. The network is simple and efficient, registering multiple meshes within seconds. At its core lies a topology-aware encoder--decoder design. Specifically, we first introduce a topology-aware point representation that encodes the anchor (reference) mesh's topology into its per-vertex features. This representation strengthens the network's understanding of the anchor-mesh geometry and disambiguates points that are Euclidean-close yet geodesically distant. We then propose a multi-modal encoder that fuses this anchor-mesh representation with complementary cues from each frame, such as shape latents and image features. These multi-source signals are compressed into a compact global motion embedding that captures dense inter-frame correspondence. A lightweight decoder then queries this global embedding with the anchor-mesh point representation, retrieving per-vertex deformations at target timestamps. Through extensive experiments across diverse motions and object categories, we show that \awesomemesh achieves state-of-the-art results on non-rigid registration. In addition, we find that our global embedding-then-query paradigm naturally enables the network to generate deformations at intermediate timestamps, which extends \awesomemesh to motion interpolation and mesh morphing. Project page: \url{https://meshloom.github.io/}.
\end{abstract}


\begin{teaserfigure}
  \centering
  \includegraphics[width=\textwidth]{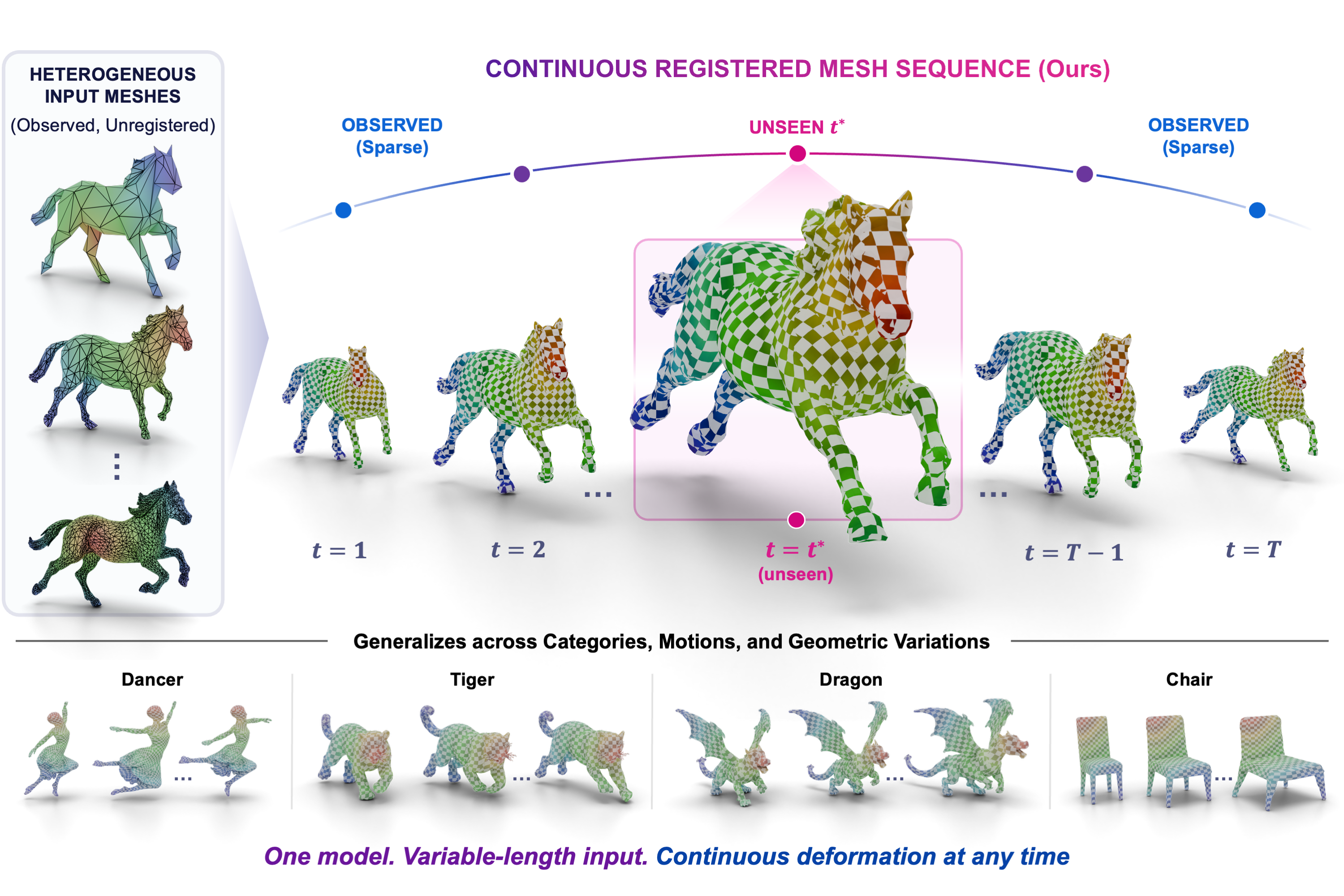}
  \caption{
  \textbf{\awesomemesh unifies input meshes into a topology-consistent output sequence.} Beyond registering the observed input frames, our network can also synthesize mesh deformations at arbitrary unseen intermediate timestamps. The method generalizes across diverse object categories and motions, and remains robust under geometric variations. The registered meshes are rendered using the same colored coordinates as the anchor mesh for visually apparent vertex correspondences across frames.
  } 
  \label{fig:teaser}
\end{teaserfigure}

\maketitle

\section{Introduction}

Non-rigid mesh registration~\cite{deng2022survey}, the task of recovering a continuous deformation field that aligns a source shape to one or more target shapes, is a long-standing problem in computer graphics and computer vision. Producing a registered mesh sequence with a single, persistent topology is a fundamental substrate for many downstream applications, including motion capture, animation retargeting, and physically-based simulation. More recently, it has also become the missing link for extending modern 3D foundation models~\cite{zhao2025hunyuan3d, xiang2025native} from static 3D reconstruction to dynamic 4D reconstruction, in which the reconstructed 3D shapes vary in vertex count and connectivity across frames. Despite decades of progress, however, an off-the-shelf registration solution that is simultaneously fast, accurate, and broadly applicable across object categories remains elusive.

Traditional registration methods rely on spatial deformation optimization~\cite{li2009robust, chang2011global, jian2005robust} or intrinsic descriptor matching~\cite{bronstein2006generalized, lipman2009mobius}. However, they are often slow and highly sensitive to initialization and topological variations, making them impractical for modern dynamic-content pipelines. Recent learning-based registration methods have emerged to address these issues by leveraging the representational power of neural networks. Yet, these methods are typically bottlenecked along one or more of the following axes. \emph{(i) Domain-restricted} models~\cite{groueix20183d, marin2024nicp, gao2026rino} do not reliably generalize to open-vocabulary objects. \emph{(ii) Iterative} models~\cite{li2022non, feng2023differentiable, jiang2023non} still require test-time optimization or repeated refinement passes, sacrificing the speed advantage of feed-forward inference. \emph{(iii) Post-processing-dependent} models~\cite{aigerman2022neural, sundararaman2022reduced, sun2023spatially} predict only intermediate Jacobian fields or correspondence maps, and require further processing to obtain vertex deformations. \emph{(iv) Pairwise} models~\cite{trappolini2021shape, sundararaman2024deformation, cao2024spectral} register only a single source--target pair at a time, and cannot ingest variable-length sequences in one pass. Crucially, no prior approach addresses all of these limitations within a single unified framework.

To close this gap, we propose \awesomemesh, a generalizable feed-forward framework that simultaneously resolves all of the above limitations within a single model.  Our approach is based on a global embedding–then–query paradigm: we encode an entire mesh sequence into a compact motion representation, which is then queried using a fixed reference mesh to recover dense deformations. Architecturally, we design \awesomemesh as a topology-aware encoder--decoder network. We first introduce a topology-aware point representation for the anchor (\ie, the reference mesh that defines the canonical vertex set) mesh, enabling the network to better extract and comprehend its geometry. Specifically, by augmenting each point's position with features aggregated over its mesh neighborhood, we embed the anchor mesh's topology into per-point descriptors. This disambiguates Euclidean-close yet geodesically distant vertices and prevents them from erroneously moving together under deformation. 
Then an encoder fuses the anchor-mesh representation with complementary cues from each mesh frame across the sequence, such as shape latents and image features. These multi-modal signals are projected into a compact global motion embedding that captures dense inter-frame correspondence. A lightweight deformation decoder then queries this embedding with the anchor-mesh point representation, retrieving per-vertex deformations at target timestamps.

Extensive evaluation across diverse object categories and motion patterns demonstrates that \awesomemesh is effective, efficient, and broadly generalizable (Fig.~\ref{fig:teaser} shows selected results). Even compared with the most recent feed-forward attempt, ActionMesh~\cite{sabathier2026actionmesh}, which takes an encouraging step toward open-vocabulary registration, our network exhibits superior performance and stronger robustness. Furthermore, we show that our global embedding-then-query paradigm naturally enables the network to predict deformations at intermediate timestamps, readily extending the method to applications such as motion interpolation and mesh morphing.

Our main contributions are summarized as follows.

\begin{itemize}[leftmargin=*, itemsep=2pt, topsep=2pt, parsep=0pt, partopsep=0pt]
    \item We propose \awesomemesh, a feed-forward non-rigid mesh registration framework that, within a single model, supports variable-length inputs, predicts explicit per-vertex deformations, and generalizes across open-vocabulary object categories, thereby resolving several bottlenecks of existing non-rigid registration methods.

    \item We introduce a topology-aware point representation that augments each point with features aggregated over its mesh neighborhood. By embedding topology information into per-point descriptors, the representation disambiguates Euclidean-close yet geodesically distant points and alleviates vertex-entanglement artifacts during deformation.

    \item We introduce a global embedding-then-query paradigm for non-rigid registration. By encoding the entire mesh sequence into a global embedding, the network can be queried at any timestamp for the corresponding vertex deformation, naturally extending \awesomemesh to motion interpolation and mesh morphing.

\end{itemize}

\begin{figure*}[!t]
  \centering
   \includegraphics[width=\linewidth]{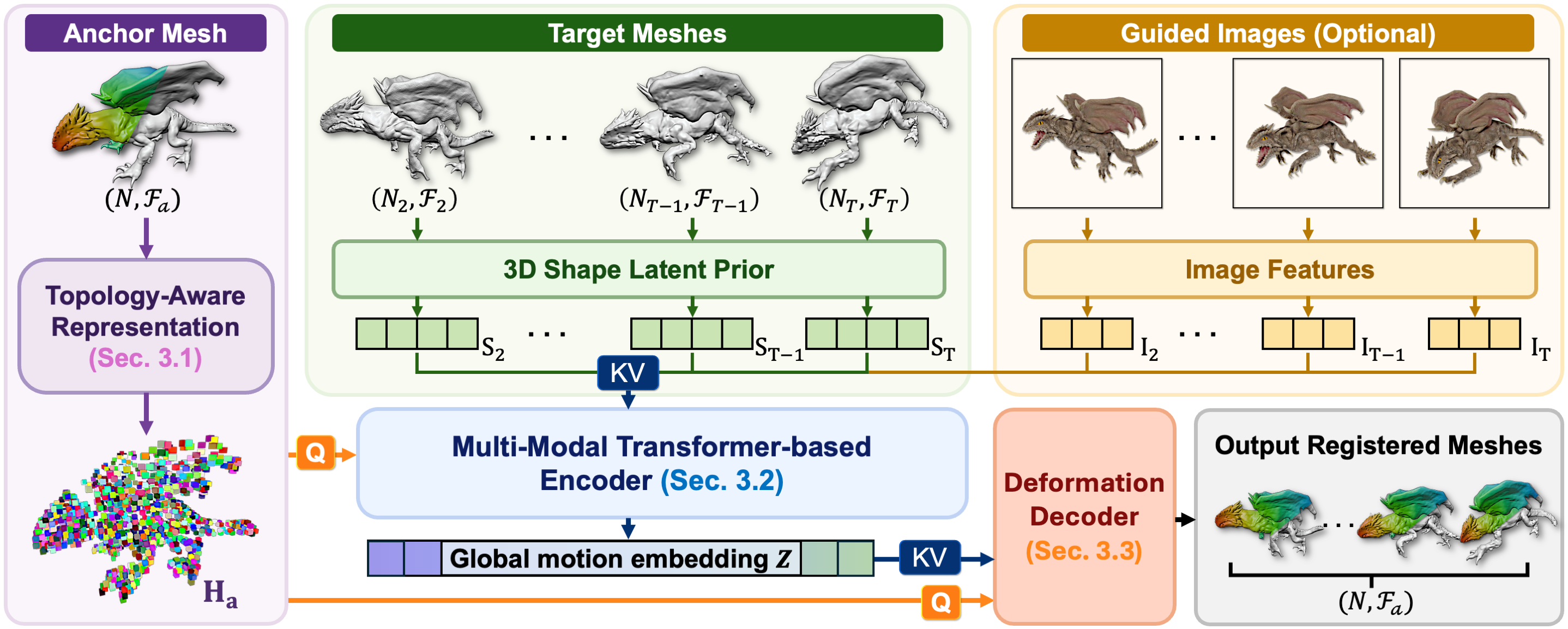}
   \caption{\textbf{Workflow of \awesomemesh.} Given an input mesh sequence whose frames differ in vertex count $N_t$ and connectivity $\mathcal{F}_t$, together with (optional) accompanying reference images, our network proceeds in three steps. (1) We designate the first frame as the anchor mesh and embed it into a topology-aware representation $\mathbf{H}_a$. (2) The remaining frames and their images are encoded into per-frame shape latents $\mathbf{S}_t$ and image features $\mathbf{I}_t$ (the anchor frame is encoded in the same way; omitted here for brevity), and all features are fused with $\mathbf{H}_a$ by a transformer-based encoder to produce a global motion embedding $\mathbf{Z}$. (3) A lightweight decoder then queries $\mathbf{Z}$ with $\mathbf{H}_a$ to predict the per-vertex deformation of the anchor mesh at every frame, yielding an output sequence with a consistent vertex count $N$ and face connectivity $\mathcal{F}_a$ across all frames.}
   \label{fig:framework}
\end{figure*}

\section{Related Work}

\paragraph{Classical non-rigid registration.} Unlike rigid registration, which estimates a single global rotation and translation, non-rigid registration aims to recover a deformation field that aligns a source surface to a target. Common deformation parameterizations include per-vertex displacements or local affine transforms~\cite{huang2011global, allen2003space}, deformation graphs~\cite{sumner2007embedded}, and Gaussian kernel-based fields~\cite{myronenko2006non}. Classical methods can be roughly split into two complementary families. \emph{Extrinsic} methods~\cite{li2009robust, chang2011global, jian2005robust} optimize the deformation directly in the ambient 3D space, typically minimizing point-to-point, point-to-plane, or probabilistic distances together with regularizers that enforce smoothness and local rigidity. \emph{Intrinsic} methods~\cite{ovsjanikov2012functional, solomon2012soft, rustamov2013map} instead exploit surface-intrinsic properties and focus on correspondence estimation rather than explicit deformation. Representative examples include minimum-distortion approaches~\cite{rodola2012game, anguelov2004correlated} that preserve geodesic or conformal structure, and spectral or functional methods~\cite{kovnatsky2013coupled, cosmo2016matching} that represent mappings in a low-frequency basis. The two families have complementary failure modes: extrinsic methods are sensitive to initialization and prone to local minima, while intrinsic methods are vulnerable to topology change and noise. Hybrid schemes~\cite{marin2020farm, eisenberger2020smooth} bridge the gap by using intrinsic techniques for initialization followed by extrinsic refinement, yet remain optimization-based and therefore slow.

\paragraph{Learning-based non-rigid registration.} To address the limitations of the classical methods, recent work has turned to neural networks for non-rigid registration. By replacing individual stages of the classical pipeline with learned counterparts, such as learned shape descriptors, learned functional maps, or networks that directly regress deformation, these methods have achieved encouraging progress. They remain, however, bottlenecked along one or more orthogonal axes. \emph{Domain-restricted} models~\cite{groueix20183d, marin2024nicp, gao2026rino} learn category-specific shape spaces, typically of humans or animals, and do not well generalize to in-the-wild objects. \emph{Iterative} models~\cite{li2022non, feng2023differentiable, jiang2023non, chen2025v2m4} embed the network inside a refinement loop or rely on per-instance test-time fitting, sacrificing the speed advantage of feed-forward inference. \emph{Post-processing-dependent} models~\cite{aigerman2022neural, sundararaman2022reduced, sun2023spatially} predict intermediate quantities such as per-vertex Jacobians or functional maps, which must then be lifted to a deformation by an external solver. \emph{Pairwise} models~\cite{trappolini2021shape, sundararaman2024deformation, cao2024spectral} register only a single source--target pair, leaving multi-frame consistency as a downstream concern. Together, these bottlenecks prevent existing learning-based methods from meeting the demands of modern dynamic-content pipelines. In contrast, we propose a feed-forward network that addresses all of these issues in a unified, efficient, and robust manner. The most direct feed-forward peer is the recent ActionMesh~\cite{sabathier2026actionmesh}, which trains a registration network and confirms the viability of feed-forward registration at scale, but remains limited to relatively simple meshes and breaks down under challenging cases. Compared with ActionMesh, our network achieves higher performance and stronger robustness across diverse scenarios.

\section{Method}
\label{sec:method}

Let $\mathcal{M} = \{\mathcal{M}_t\}_{t=1}^{T}$ denote an input sequence of $T$ meshes, where $\mathcal{M}_t = (\mathcal{V}_t, \mathcal{F}_t)$ has vertices $\mathcal{V}_t \in \mathbb{R}^{N_t \times 3}$ and faces $\mathcal{F}_t$. Both the vertex count and the face connectivity may differ across frames, \ie, $N_{t_1} \neq N_{t_2}$ or $\mathcal{F}_{t_1} \neq \mathcal{F}_{t_2}$ for two distinct frame indices ${t_1}, {t_2}$. Without loss of generality, we designate the first frame as the \emph{anchor} mesh $\mathcal{M}_{a} = (\mathcal{V}_{a}, \mathcal{F}_{a}) = (\mathcal{V}_{1}, \mathcal{F}_{1})$, which contains $N = N_1$ vertices and is registered to all other frames (other frames can also be selected as the anchor by reordering the sequence). The network then unifies these mesh frames into a topology-consistent output sequence:
\begin{equation}
\widetilde{\mathcal{M}} = \mathrm{Network}(\mathcal{M}) = \big\{\,(\widetilde{\mathcal{V}}_t,\,\mathcal{F}_{a})\,\big\}_{t=1}^{T},
\qquad \widetilde{\mathcal{V}}_t \in \mathbb{R}^{N \times 3}
\end{equation}
in which every frame shares the anchor topology $\mathcal{F}_{a}$ and the same number of vertices $N$.

Fig.~\ref{fig:framework} illustrates the overall pipeline, which is built on a topology-aware encoder--decoder design. First, we introduce a \emph{topology-aware point representation} (Sec.~\ref{sec:topo}) for the anchor mesh. The multi-modal \emph{motion encoder} (Sec.~\ref{sec:encoder}) then fuses this anchor-mesh representation with multi-source cues across all $T$ mesh frames, producing a global motion embedding $\mathbf{Z}$ that implicitly encodes inter-frame correspondence. The \emph{deformation decoder} (Sec.~\ref{sec:decoder}) queries $\mathbf{Z}$ with the anchor-mesh representation and predicts the per-vertex deformation at each frame.

\subsection{Topology-Aware Point Representation}
\label{sec:topo}

Since the anchor mesh defines the topology of the output sequence, with every output frame inheriting its vertex count and face connectivity, it is crucial for the network to faithfully understand its underlying shape and topology.

\begin{figure}[!t]
  \centering
  \includegraphics[width=0.9\linewidth]{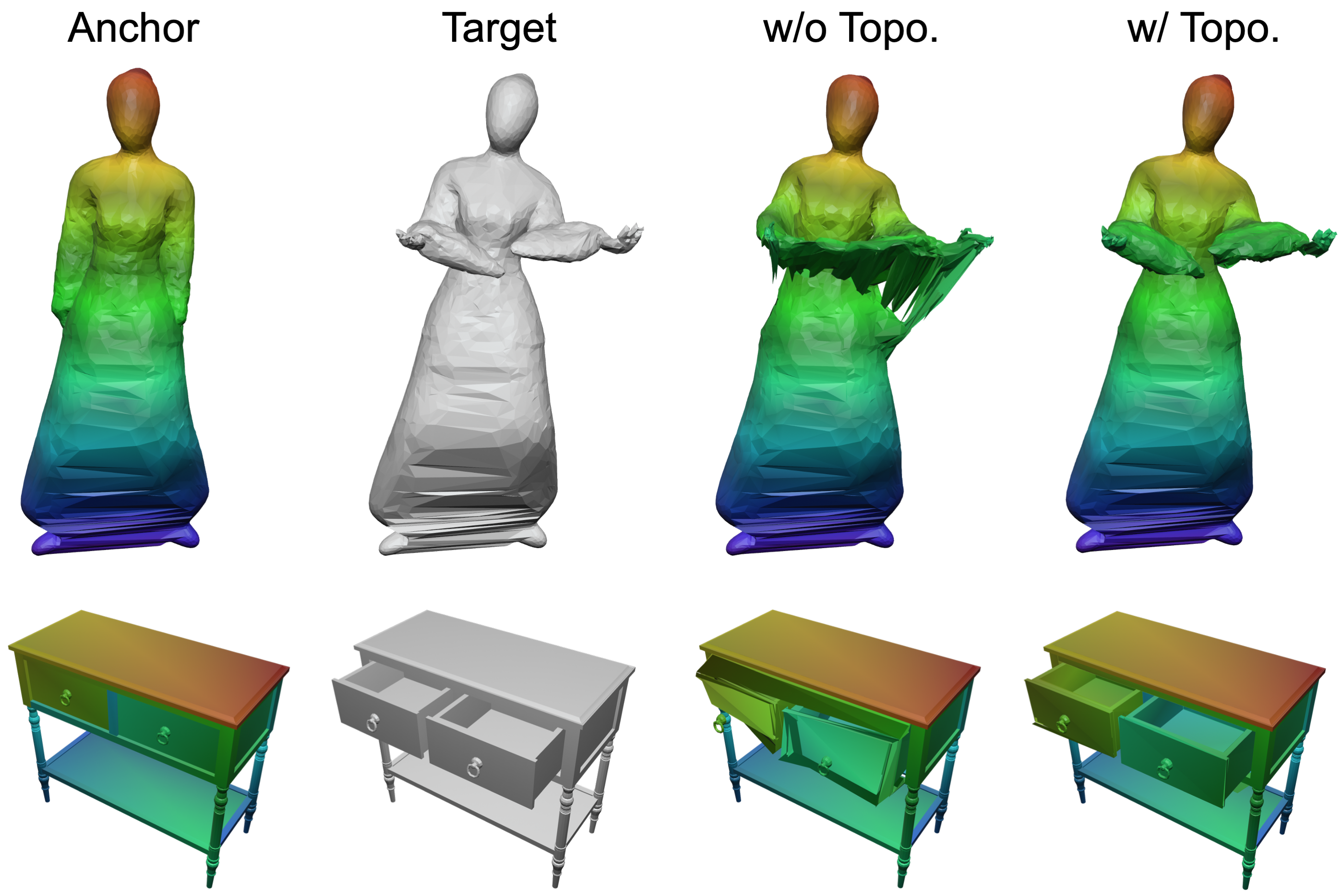}
  \caption{\textbf{Effect of the topology-aware point representation.} Qualitative comparison with and without topology information embedded in the per-vertex features. From left to right are the anchor mesh, the target mesh, and the registered results of each variant.}
  \label{fig:topo}
\end{figure}

Mainstream networks~\cite{zhang20233dshape2vecset, xiang2025structured, xiang2025native} encode only vertex positions and surface normals~\cite{sabathier2026actionmesh}, ignoring the mesh connectivity. Consequently, they fail to distinguish two vertices that are Euclidean-close but topologically distant, \eg, sleeve--torso contact region (see Fig.~\ref{fig:topo}). It is therefore prone to vertex entanglement, in which such close-but-disjoint vertices erroneously move together during deformation.

To resolve this ambiguity, we employ a GCN-style~\cite{kipf2017semisupervised} network to extract topology-aware features for each vertex. The network takes a vertex feature matrix $\mathbf{H} \in \mathbb{R}^{N\times d}$ as input and aggregates mesh connectivity by propagating vertex features across the graph structure. The feature matrix $\mathbf{H}$ is constructed by concatenating the position-embeddings and surface normals of the mesh. To implement the aggregation, we define a row-normalized adjacency operator $\mathbf{A}\in\mathbb{R}^{N\times N}$ induced by the face set $\mathcal{F}_{a}$:
\begin{equation}
    {A}_{ij} =
    \begin{cases}
        1/\deg(i), & \text{if } (i,j)\in\mathcal{E},\\
        0, & \text{otherwise.}
    \end{cases}
\end{equation}
Here, $\mathcal{E}=\{(i,j)\} $ denotes the edge set of the input mesh, and $\deg(i)$ is the degree of vertex $i$. Therefore, the $i$-th row of $\mathbf{A}\mathbf{H}$ represents the feature average across the one-ring neighborhood of vertex $i$. 
To effectively capture the overall topology, we enlarge the receptive field by applying the normalized adjacency operator multiple times. This yields $\mathbf{A}^{p}\mathbf{H}$, which aggregates information from $p$-hop neighborhoods. We then stack $L$ residual blocks on the mesh graph, with each layer defined as:
\begin{equation}
\label{eq:gcn}
    \mathbf{H}^{(\ell+1)} = \mathrm{LayerNorm}\!\left(
        \mathbf{H}^{(\ell)} +
        \mathrm{MLP}\!\left(
            \left[\mathbf{H}^{(\ell)} \,\middle\|\,
            \mathbf{A}^{p}\mathbf{H}^{(\ell)}\right]
        \right)
    \right)
\end{equation}
where $[\cdot\|\cdot]$ denotes channel-wise concatenation. We define the final layer output $\mathbf{H}^{L}$ as the topology-aware representation. For anchor mesh, we denote its corresponding feature matrix as $\mathbf{H}_{a}\in\mathbb{R}^{N\times d}$.

\subsection{Motion Encoder}
\label{sec:encoder}

The motion encoder aims to build a global embedding that describes the input mesh sequence and captures inter-frame correspondence. To enrich the input features and help the network better comprehend the sequence, the encoder ingests features from multiple sources: both the topology-aware representation of the anchor mesh $\mathbf{H}_{a}$ and per-frame features across the sequence, such as 3D shape-latent priors and image features. These features are processed by several transformer blocks to produce the global motion embedding $\mathbf{Z}$.

\paragraph{Encoder Input.} The encoder takes topology-aware representations as its input. A natural design is to use topology-aware features from each individual frame. However, this requires running the GCN-style network for every frame, resulting in substantial computational overhead. Empirically, we find that using only the anchor-mesh representation achieves comparable performance while significantly reducing computation (see Table~\ref{tab:ablate input}). Therefore, we use only $\mathbf{H}_{a}$ as the encoder input. Specifically, since the original feature matrix is dense, we first downsample $\mathbf{H}_{a}$ to $N_{\mathrm{ds}}$ points using farthest-point sampling (FPS), obtaining $\mathbf{H}_{a}^{\mathrm{ds}} \in \mathbb{R}^{N_{\mathrm{ds}}\times d}$. We then replicate $\mathbf{H}_{a}^{\mathrm{ds}}$ across all $T$ time steps to form the initial encoder input $\mathbf{Z}^{(0)}\in\mathbb{R}^{T\times N_{\mathrm{ds}}\times d}$. This design improves efficiency while preserving performance.

\paragraph{Multi-Modal Transformer Block.} Each transformer block comprises two cross-attention layers followed by two self-attention layers. 
The two cross-attention layers attend to two complementary sources, respectively. The first is a per-frame \emph{3D shape-latent prior} $\{\mathbf{S}_{t}\}_{t=1}^{T}$, where each $\mathbf{S}_{t}\in\mathbb{R}^{N_{s}\times d}$ is the latent of frame $t$ produced by an off-the-shelf pretrained 3D foundation model (the size of $\mathbf{S}_{t}$ may vary across frames, we omit this dependency in the notation for clarity). Having been trained on large-scale 3D-asset datasets, $\mathbf{S}_{t}$ encodes strong priors over mesh geometry that sharpen the encoder's understanding of per-frame shape structure. In addition, we optionally allow the network to leverage per-frame \emph{image features} $\{\mathbf{I}_{t}\}_{t=1}^{T}$, where $\mathbf{I}_{t}\in\mathbb{R}^{N_{m}\times d}$ is extracted from a reference image of frame $t$ and carries object-appearance cues that further enrich the semantic representation. This optional conditioning is highly compatible with modern Image-to-3D pipelines, where reference images are readily available. Formally, given the input tokens $\mathbf{Z}_{t} \in \mathbb{R}^{N_{\mathrm{ds}}\times d}$, the feature injection process is defined as:
\begin{equation}
\begin{split}
\mathbf{Z}_{t}^{s} = \mathrm{CrossAttn}\!\big(\mathrm{Q}\!\leftarrow\!\mathbf{Z}_{t},\,\mathrm{KV}\!\leftarrow\!\mathbf{S}_{t}\big)\\
\mathbf{Z}_{t}^{s,m} = \mathrm{CrossAttn}\!\big(\mathrm{Q}\!\leftarrow\!\mathbf{Z}_{t}^{s},\,\mathrm{KV}\!\leftarrow\!\mathbf{I}_{t}\big)
\end{split}
\end{equation}
where the $\mathrm{Q}/\mathrm{KV}$ tags indicate which input provides the queries and which provides the keys and values, and $\mathbf{Z}_{t}^{s}$ and $\mathbf{Z}_{t}^{s,m}$ represent the  features enriched by the 3D shape prior and image guidance.

The self-attention layers consist of an \emph{inter-frame} self-attention, which is applied to the concatenation of tokens from all $T$ frames to fuse temporal information, and an \emph{intra-frame} self-attention, which is applied independently within each frame to refine spatial structure and normalize per-frame activations. Letting $\mathbf{Z}_{1:T}\in\mathbb{R}^{(T\cdot N_{\mathrm{ds}})\times d}$ denote the concatenation of all per-frame tokens at the current layer, the self-attention layers can be formulated as:
\begin{equation}
\mathrm{SelfAttn}(\mathrm{QKV}\!\leftarrow\!\mathbf{X}),\qquad
\mathbf{X}=
\begin{cases}
\mathbf{Z}_{1:T} & \text{(inter-frame)},\\
\mathbf{Z}_{t}   & \text{(intra-frame)}.
\end{cases}
\end{equation}

After multiple layers of transformer blocks, we concatenate all output tokens across the temporal dimension to form the final global motion embedding $\mathbf{Z}\in\mathbb{R}^{(T\cdot N_{\mathrm{ds}})\times d}$.

\subsection{Deformation Decoder}
\label{sec:decoder}

\begin{figure}[!t]
  \centering
  \includegraphics[width=\linewidth]{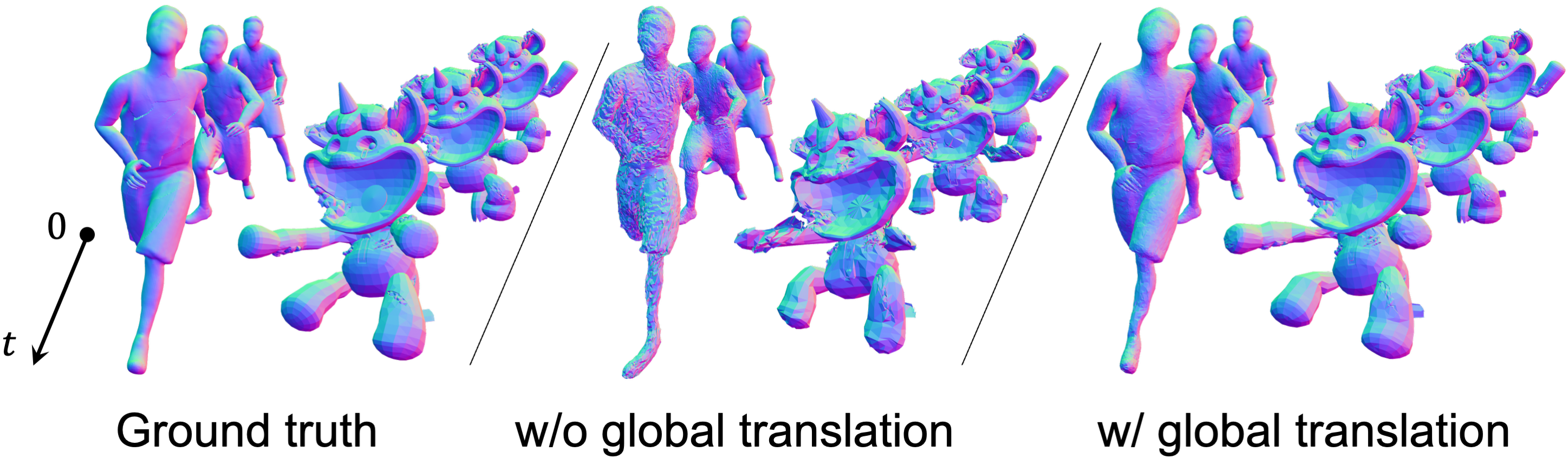}
  \caption{\textbf{Effect of the global-translation branch.} From left to right are the input mesh frames, the registered results from a single-branch decoder, and the registered results from our two-branch decoder with an additional global-translation prediction. Zoom in on the mesh surfaces to compare smoothness and structural stability.}
  \label{fig:global}
\end{figure}

Given the global embedding $\mathbf{Z}$, the decoder queries it to obtain per-vertex deformations of the anchor mesh at any timestep, yielding a mesh sequence with consistent topology. Specifically, for a anchor vertex $\mathbf{v}_{i}\in\mathbb{R}^{3}$ ($i\in\{1,\dots,N\}$) and a target timestamp $t$, the decoder predicts a deformed position $\widetilde{\mathbf{v}}_{i}^{(t)}\in\mathbb{R}^{3}$. Rather than regressing the full displacement directly, we decompose it into two components: a \emph{global centroid shift} $\Delta\mathbf{c}^{(t)}\in\mathbb{R}^{3}$ and a \emph{local residual} $\Delta\mathbf{r}_{i}^{(t)}\in\mathbb{R}^{3}$:
\begin{equation}
\widetilde{\mathbf{v}}_{i}^{(t)} = \mathbf{v}_{i} + \Delta\mathbf{c}^{(t)} + \Delta\mathbf{r}_{i}^{(t)}
\end{equation}
Specifically, the global centroid shift denotes the rigid translation between the mesh centroid at frame $t$. The local residual describes the per-vertex non-rigid deformation in the local frame. Compared with a single-branch design, we empirically find that this decoupling improves robustness under large translational motion and suppresses high-frequency surface artifacts (see Fig.~\ref{fig:global}).

\paragraph{Global-translation branch.} To estimate the global translation $\Delta\mathbf{c}^{(t)}$ at any given time $t$, we first apply average pooling to $\mathbf{Z}$ across all $T\cdot N_{\mathrm{ds}}$ tokens. This yields a single sequence-level descriptor $\bar{\mathbf{z}}\in\mathbb{R}^{d}$, which encodes the global motion trajectory and provides a consistent context for the entire animation. To resolve the specific translation for a target timestamp $t$, we concatenate $\bar{\mathbf{z}}$ with the Fourier time embedding $\phi(t)\in\mathbb{R}^{d_\phi}$ and predict the centroid shift through a lightweight MLP:
\begin{equation}
\Delta\mathbf{c}^{(t)} = \mathrm{MLP}\!\big([\bar{\mathbf{z}}\,\|\,\phi(t)]\big)
\end{equation}

\paragraph{Local-deformation branch.} To predict the local residual $\Delta\mathbf{r}_{i}^{(t)}$, we employ a single cross-attention block. Specifically, for each vertex $i$, we take its topology representation $\mathbf{h}_{i}\in\mathbb{R}^{d}$  ($i$-th row of $\mathbf{H}_{a}$) and modulate it with the target timestamp $t$ via FiLM~\cite{perez2018film} as the query. We then use the global motion embedding $\mathbf{Z}$ as the keys and values. The cross-attention output is then projected by an MLP to yield the local residual:
\begin{equation}
\label{eq:local_deform}
\Delta\mathbf{r}_{i}^{(t)} = \mathrm{MLP} \!\big(\mathrm{CrossAttn}\!\big(\mathrm{Q}\!\leftarrow\!\mathrm{FiLM}(\mathbf{h}_{i}, t),\,\mathrm{KV}\!\leftarrow\!\mathbf{Z}\big)\big)
\end{equation}

Since both the global and local components at the target timestamp are decoded directly from $\mathbf{Z}$ conditioned solely on $t$, the decoder can naturally be queried at arbitrary timestamps that the encoder has never observed. This property allows \awesomemesh to extend beyond registration to applications such as motion interpolation and mesh morphing, as we demonstrate in Sec.~\ref{subsec: additional res}.

\begin{figure}[!t]
  \centering
  \includegraphics[width=\linewidth]{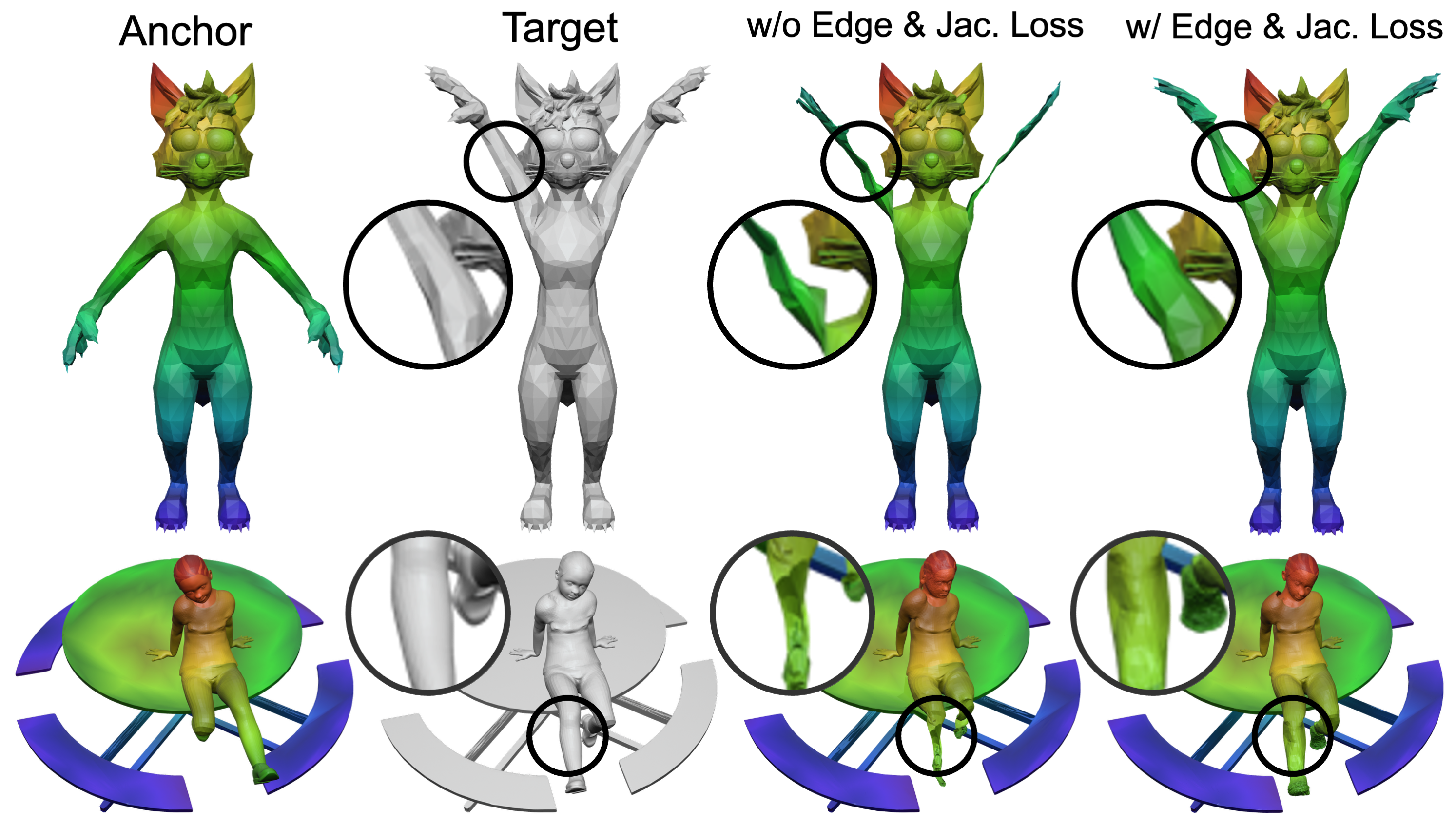}
  \caption{\textbf{Effect of the edge-length and Jacobian losses.} Qualitative comparison under different loss configurations. From left to right are the anchor mesh, the target mesh, and the registered results. We also provide zoomed-in views for clearer visualization of the differences.}
  \label{fig:loss}
\end{figure}

\begin{table*}[t]
\caption{\textbf{Performance comparison on ActionBench.} We report both geometric metrics (CD-3D, CD-4D, CD-Motion) and rendering-based visual metrics (CLIP, LPIPS, DreamSim). Each score is averaged over the $2$-, $4$-, and $8$-frame input settings; per-setting results are provided in Appendix~\ref{app:results}. We also report inference time, measured on $8$-frame inputs with a fixed $150$K vertices and $300$K faces per mesh frame. Best results are in \textbf{bold}, second-best are \underline{underlined}.}
\label{tab:compare}
\centering
\resizebox{1\linewidth}{!}{
\setlength{\tabcolsep}{15pt}
\begin{tabular}{@{}l|cccccc|c@{}}
\toprule
\multicolumn{1}{l|}{\textbf{Method}} & \textbf{CD-3D   $\downarrow$} & \textbf{CD-4D   $\downarrow$} & \textbf{CD-M   $\downarrow$} & \textbf{CLIP   $\uparrow$} & \textbf{LPIPS   $\downarrow$} & \textbf{DreamSim   $\downarrow$} & Inf. Time (s) \\ \midrule
TransferMatch~\cite{trappolini2021shape} & 0.0975 & 0.1762 & 0.8457 & 0.7610 & 0.1670 & 0.2455 & 14.2 \\
NDP~\cite{li2022non} & {\ul 0.0545} & 0.0843 & 0.1699 & 0.9666 & 0.0623 & 0.0325 & {\ul 5.9} \\
LNDP~\cite{li2022non} & 0.0585 & 0.0852 & 0.1760 & 0.9605 & 0.1037 & 0.0480 & 279.9 \\
DPF~\cite{prokudin2023dynamic} & \cellcolor[HTML]{EFEFEF}\textbf{0.0531} & \cellcolor[HTML]{EFEFEF}\textbf{0.0820} & 0.1770 & 0.9715 & \cellcolor[HTML]{EFEFEF}\textbf{0.0369} & 0.0208 & 70.2 \\
ClusterReg~\cite{zhao2024correspondence} & {\ul 0.0545} & 0.0835 & 0.1902 & 0.8752 & 0.1130 & 0.1045 & 42.6 \\
OAReg~\cite{zhao2025occlusion} & 0.0562 & 0.0847 & 0.1592 & 0.9675 & 0.0641 & 0.0334 & 25.3 \\
ActionMesh~\cite{sabathier2026actionmesh} & 0.0560 & 0.0842 & 0.1557 & 0.9707 & 0.0821 & 0.0287 & 7.7 \\ \midrule
Ours (w/o img) & 0.0571 & 0.0840 & {\ul 0.1493} & {\ul 0.9786} & 0.0408 & {\ul 0.0197} & \cellcolor[HTML]{EFEFEF} \\
Ours & 0.0567 & {\ul 0.0834} & \cellcolor[HTML]{EFEFEF}\textbf{0.1471} & \cellcolor[HTML]{EFEFEF}\textbf{0.9793} & {\ul 0.0399} & \cellcolor[HTML]{EFEFEF}\textbf{0.0184} & \multirow{-2}{*}{\cellcolor[HTML]{EFEFEF}\textbf{3.1}} \\ \bottomrule
\end{tabular}
}
\end{table*}

\begin{figure*}[!t]
  \centering
  \includegraphics[width=0.95\linewidth]{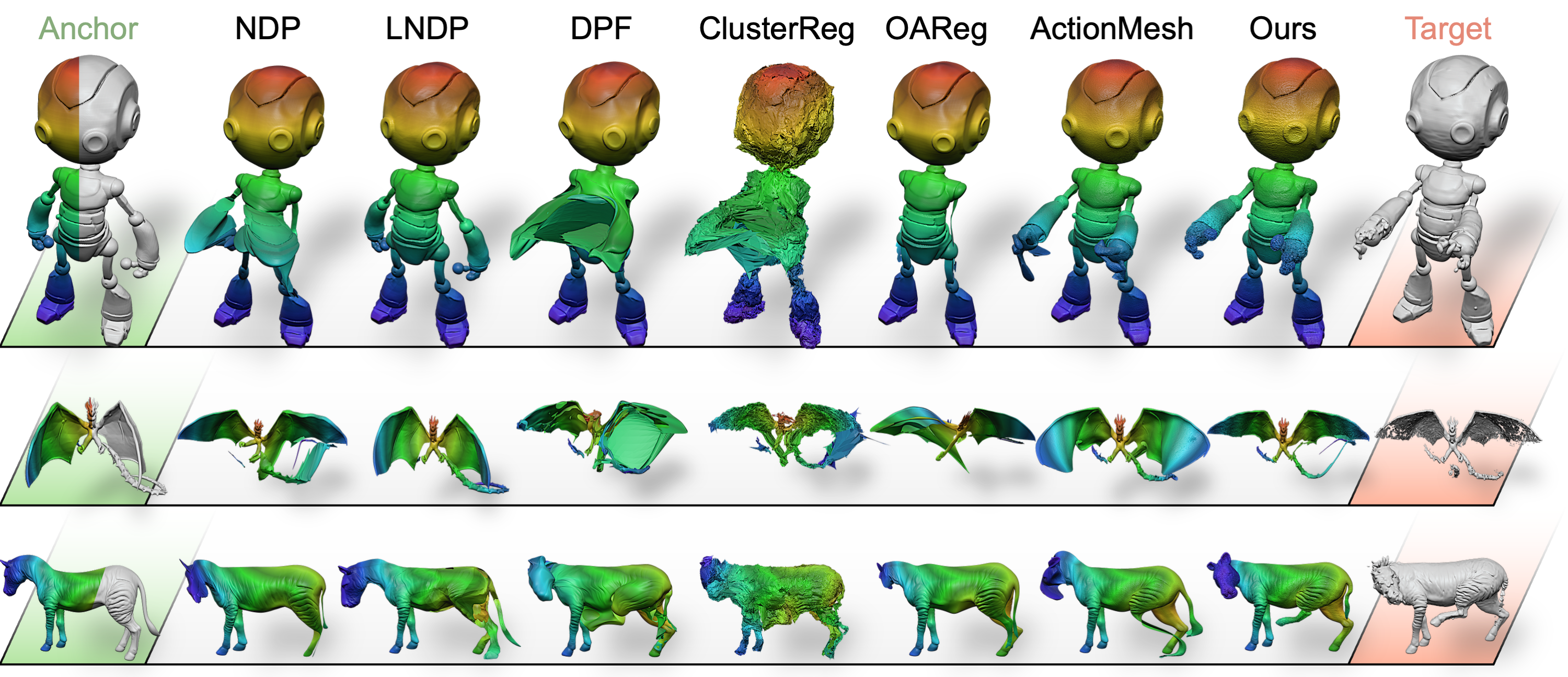}
  \caption{\textbf{Visual comparison with baseline methods.} From left to right: the anchor mesh, the registered results of each method, and the target mesh. We render each registered mesh using the same colored coordinates as the anchor, so that vertex correspondences across frames are visually apparent.}
  \label{fig:compare}
\end{figure*}

\begin{figure*}[!t]
  \centering
  \includegraphics[width=\linewidth]{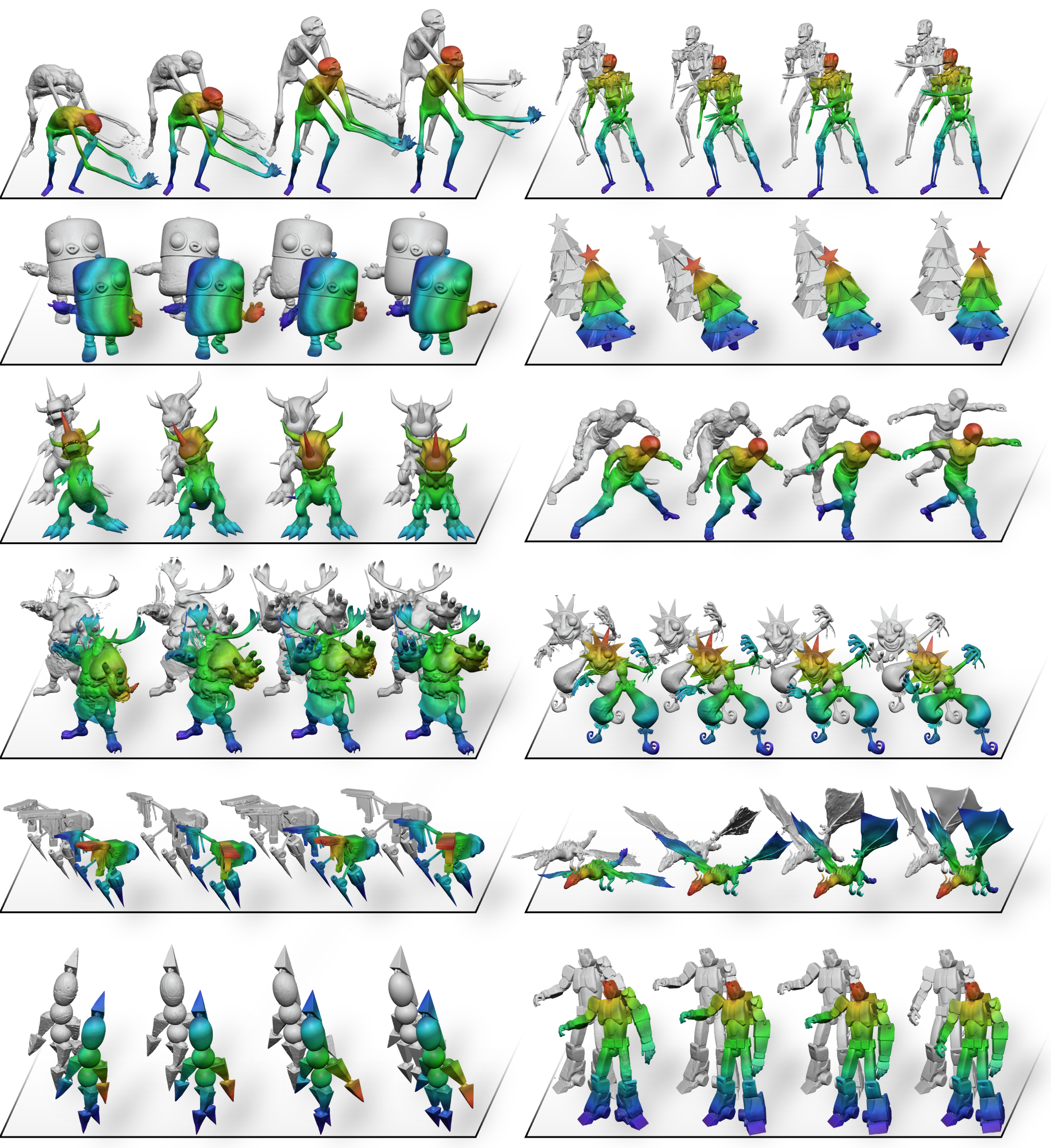}
  \caption{\textbf{Additional registration results of our method.} We show our registration results across a broader range of motions and object identities. Unregistered input meshes are rendered in pure white, while registered outputs are rendered with colored coordinates that visualize the vertex correspondence across mesh frames.}
  \label{fig:compare3}
\end{figure*}

\begin{figure*}[!t]
  \centering
  \includegraphics[width=0.95\linewidth]{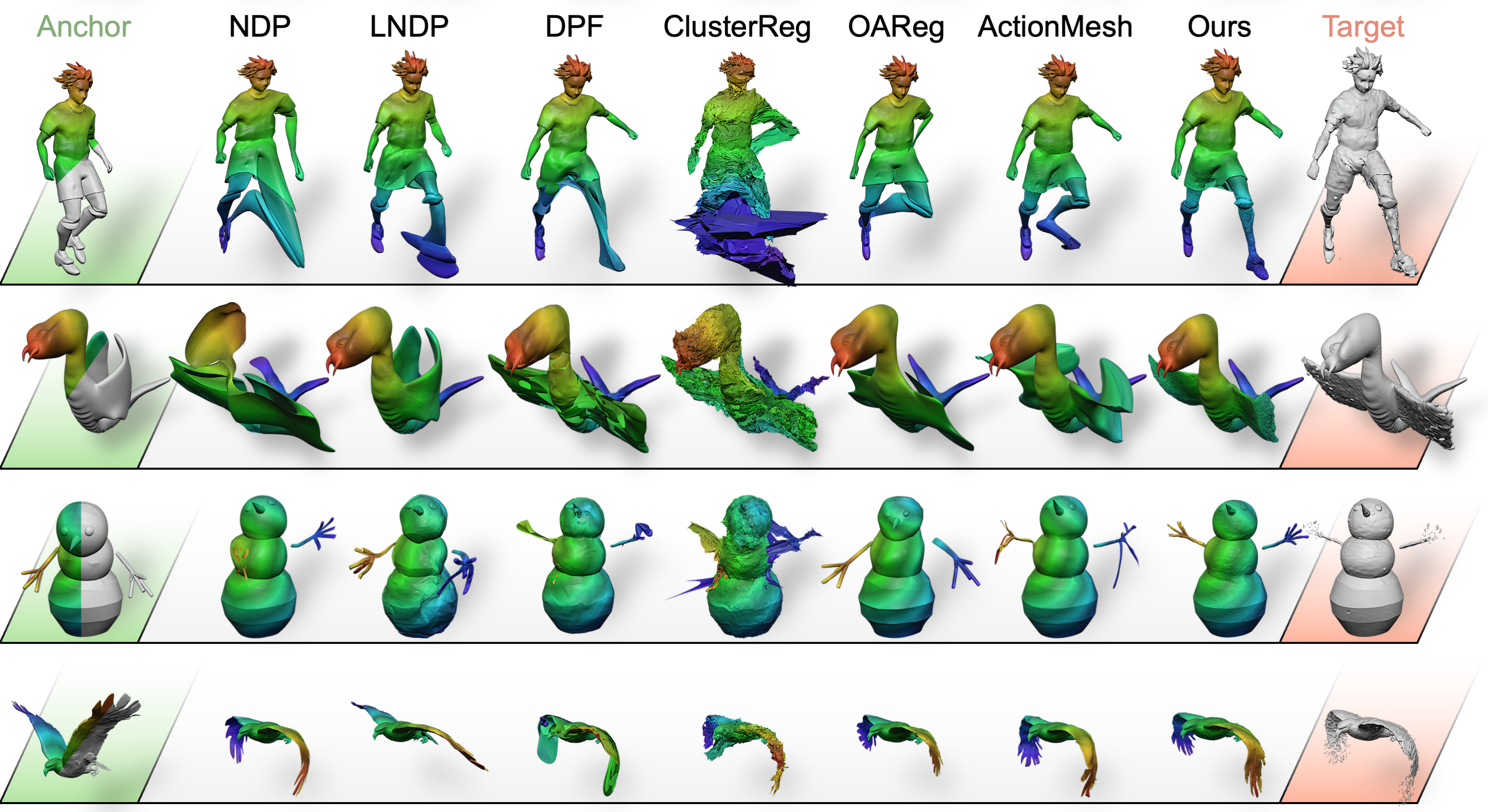}
  \caption{\textbf{Additional visual comparisons with baseline methods.} From left to right: the anchor mesh, the registered results of each method, and the target mesh. Since all methods start from the same anchor mesh, we render each registered mesh using the same colored coordinates as the anchor, so that vertex correspondences across frames are visually apparent.}
  \label{fig:compare2}
\end{figure*}

\subsection{Training Objectives}
\label{sec:training}

We train \awesomemesh end-to-end with several objectives. The global and local deformations predicted in Sec.~\ref{sec:decoder} are directly supervised with a mean-squared error:
\begin{equation}
    \mathcal{L}_{\text{global}}
    = \bigl\| \Delta\mathbf{c}^{(t)} - \Delta\mathbf{c}^{*(t)} \bigr\|_{2}^{2},
\qquad
    \mathcal{L}_{\text{local}}
    = \frac{1}{N}\sum_{i=1}^{N}
      \bigl\| \Delta\mathbf{r}_{i}^{(t)} - \Delta\mathbf{r}_{i}^{*(t)} \bigr\|_{2}^{2}
\end{equation}
where $\Delta\mathbf{c}^{*(t)}$ and $\Delta\mathbf{r}_{i}^{*(t)}$ are the ground-truth global translation and local residual, respectively.

In addition to direct vertex supervision, we introduce two additional losses defined on the anchor-mesh connectivity graph, which we find further facilitate training and reduce artifacts such as fold-overs and volume collapse (see Fig.~\ref{fig:loss}).

The \emph{edge-length loss} regularizes the local mesh structure by minimzing the $L_{1}$ difference between the predicted edge vectors and their ground-truth counterpart:
\begin{equation}
    \mathcal{L}_{\text{edge}}
    = \frac{1}{|\mathcal{E}|}\sum_{(i,j)\in\mathcal{E}}
      \bigl\|
        \big(\widetilde{\mathbf{v}}_{i}^{(t)} - \widetilde{\mathbf{v}}_{j}^{(t)}\big)
        - \big(\mathbf{v}_{i}^{*(t)} - \mathbf{v}_{j}^{*(t)}\big)
      \bigr\|_{1}
\end{equation}
where $\mathbf{v}_{i}^{*(t)}$ is the ground-truth position of vertex $i$ at timestep $t$, and $\mathcal{E}$ is the edge set of the anchor mesh.

To better capture the local deformation field, we further adopt a \textit{Jacobian loss}, supervising a local Jacobian at each vertex. Specifically, for vertex $i$ we estimate a Jacobian matrix $\mathbf{J}_{i}\in\mathbb{R}^{3\times 3}$ that best fits the deformation of its neighborhood in a least-squares sense. A ground-truth Jacobian  $\mathbf{J}_{i}^{*}$ is analogously derived from the reference deformation. We penalize their discrepancy with a smooth-$L_{1}$ loss:
\begin{equation}
    \mathcal{L}_{\text{jac}}
    = \frac{1}{N}\sum_{i=1}^{N}
      \mathrm{SmoothL1}\!\left(\mathbf{J}_{i},\;\mathbf{J}_{i}^{*}\right)
\end{equation}
This supervision provides a more precise structural constraint than vertex-position loss alone.
Besides, unlike local-rigidity regularizers such as as-rigid-as-possible~\cite{sorkine2007rigid}, which impose a strong inductive bias by constraining local transformations to be rigid, our Jacobian supervision adopts a data-driven approach. By supervising the full deformation gradient against the target, our method preserves the fidelity of the local deformation field and naturally accommodates complex non-rigid effects, such as stretching and shearing, that are essential for realistic motion. More implementation details of our losses could be found in Appendix~\ref{app:train_obj}.

The overall training objective is:
\begin{equation}
    \mathcal{L}
    = \lambda_{\text{global}}\,\mathcal{L}_{\text{global}}
    + \lambda_{\text{local}}\,\mathcal{L}_{\text{local}}
    + \lambda_{\text{edge}}\,\mathcal{L}_{\text{edge}}
    + \lambda_{\text{jac}}\,\mathcal{L}_{\text{jac}}
\end{equation}

\section{Experiments}
\label{sec:experiments}

\paragraph{Datasets.}
We train our network on the animation subset of Texverse~\cite{zhang2025texverse}, which contains animated-object sequences spanning diverse categories such as characters, animals, and mechanical objects. After applying a series of filtering steps (detailed in Appendix~\ref{app:data}), approximately $30$K high-quality $32$-frame training sequences remain. For evaluation, we adopt ActionBench~\cite{sabathier2026actionmesh}, which covers diverse object categories and rich motions and is therefore well suited for assessing the practical performance of non-rigid registration methods in general settings.

\paragraph{Evaluation metrics.}
We evaluate along two complementary axes: \emph{geometric alignment} and \emph{rendering-based alignment}. For geometric alignment, we follow the ActionBench protocol~\cite{sabathier2026actionmesh} and report three geometric metrics. \emph{CD-3D} measures shape accuracy between the ground-truth and reconstructed deformed meshes, independent of global pose drift. \emph{CD-4D} jointly captures shape fidelity and temporal pose consistency at the object level. \emph{CD-Motion} further probes motion coherence by measuring vertex-level deformation consistency. For rendering-based alignment, we render normal maps from multiple viewpoints across all mesh frames and compare those of the ground-truth and reconstructed meshes. We adopt the widely used \emph{LPIPS}~\cite{zhang2018unreasonable} and \emph{CLIP Similarity}~\cite{radford2021learning} metrics, and additionally report \emph{DreamSim}~\cite{fu2023dreamsim}, which provides a more human-aligned perceptual distance that correlates well with global structural similarity. More information of these metrics could be found in Appendix~\ref{app:metric}.

\begin{figure*}[!t]
  \centering
  \includegraphics[width=0.95\linewidth]{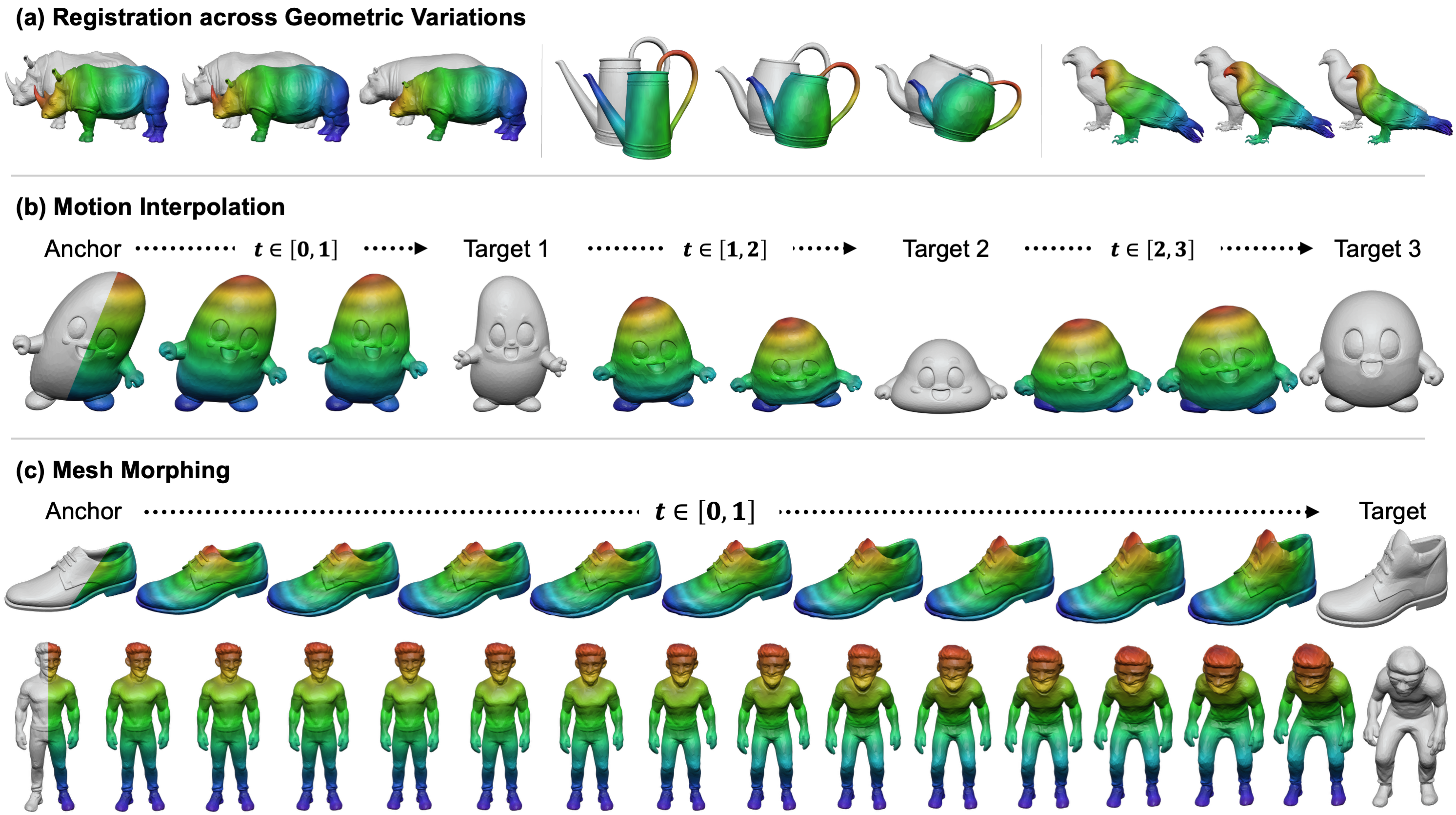}
  \caption{\textbf{Registration across geometric variations, motion interpolation, and mesh morphing.} Unregistered input meshes are rendered in pure white, while registered or interpolated outputs are rendered with colored coordinates. The anchor mesh is shown half in pure white and half in colored coordinates.}
  \label{fig:application}
\end{figure*}

\subsection{Comparisons}

We compare our method against recent state-of-the-art non-rigid registration methods. Specifically, we adopt TransferMatch~\cite{trappolini2021shape}, NDP~\cite{li2022non} and its supervised-based variant LNDP~\cite{li2022non}, DPF~\cite{prokudin2023dynamic}, ClusterReg~\cite{zhao2024correspondence}, OAReg~\cite{zhao2025occlusion}, and the recent ActionMesh~\cite{sabathier2026actionmesh} as baselines, and compare against them both quantitatively and qualitatively.

\paragraph{Quantitative comparisons.} We evaluate with three input lengths ($2$, $4$, and $8$ frames) and report the averaged performance in Table~\ref{tab:compare}; per-setting numbers are given in Appendix~\ref{app:results}. Our method ranks at the top of many metrics. In particular, we obtain the best CD-Motion, which directly measures the temporal consistency of per-vertex deformation trajectories (see Appendix~\ref{app:metric}) and is therefore the most discriminative indicator of registration quality, improving on the previous best by more than $5\%$. DPF~\cite{prokudin2023dynamic} performs well on CD-3D/4D, partly because its Chamfer-distance supervision directly matches these metrics. However, it is prone to vertex-entanglement artifacts (Fig.~\ref{fig:compare}). Our method is consistently top-ranked on visual metrics, exceeding the previous best by over $10\%$ on DreamSim, which indicates more natural deformations. Notably, our method delivers all of these gains at the fastest inference speed, roughly $50\%$ faster than the closest baseline, demonstrating that our design is both efficient and effective.

\paragraph{Qualitative comparisons.} We present qualitative comparisons in Fig.~\ref{fig:compare}; additional examples are provided in Fig.~\ref{fig:compare2} and Fig.~\ref{fig:compare3} due to space constraints. Across diverse motions and object categories, our method registers the anchor mesh to the target with noticeably fewer artifacts than the baselines.

\begin{table}[t]
\caption{\textbf{Ablation of encoder-input strategies.} We compare feeding a per-frame topology-aware representation against replicating a single anchor-mesh representation across all frames (our default). Peak training memory (GB) is also reported. Best results per metric are highlighted.}
\label{tab:ablate input}
\centering
\resizebox{\linewidth}{!}{
\begin{tabular}{@{}lccccccc@{}}
\toprule
Encoder Input & CD-3D $\downarrow$ & CD-4D $\downarrow$ & CD-M $\downarrow$ & CLIP $\uparrow$ & LPIPS $\downarrow$ & DreamSim $\downarrow$ & Mem. $\downarrow$ \\ \midrule
Per-frame & 0.0574 & 0.0857 & \cellcolor[HTML]{EFEFEF}\textbf{0.1523} & 0.9747 & 0.0487 & 0.0239 & 37.48 \\
Shared anchor & \cellcolor[HTML]{EFEFEF}\textbf{0.0573} & \cellcolor[HTML]{EFEFEF}\textbf{0.0853} & 0.1539 & \cellcolor[HTML]{EFEFEF}\textbf{0.9753} & \cellcolor[HTML]{EFEFEF}\textbf{0.0483} & \cellcolor[HTML]{EFEFEF}\textbf{0.0231} & \cellcolor[HTML]{EFEFEF}\textbf{23.54} \\ \bottomrule
\end{tabular}
}
\end{table}

\begin{table}[t]
\caption{\textbf{Ablation of encoder modalities.} We measure the contribution of the 3D shape-latent prior and the image features to overall registration performance. Best results per metric are highlighted.}
\label{tab:ablate condition}
\centering
\resizebox{\linewidth}{!}{
\begin{tabular}{@{}lcccccc@{}}
\toprule
Modality & CD-3D $\downarrow$ & CD-4D $\downarrow$ & CD-M $\downarrow$ & CLIP $\uparrow$ & LPIPS $\downarrow$ & DreamSim $\downarrow$ \\ \midrule
w/o Shape & 0.0746 & 0.1333 & 0.2518 & 0.9542 & 0.1030 & 0.0577 \\
w/o Image & 0.0577 & 0.0855 & \cellcolor[HTML]{EFEFEF}\textbf{0.1522} & 0.9740 & 0.0503 & 0.0248 \\
Full & \cellcolor[HTML]{EFEFEF}\textbf{0.0573} & \cellcolor[HTML]{EFEFEF}\textbf{0.0853} & 0.1539 & \cellcolor[HTML]{EFEFEF}\textbf{0.9753} & \cellcolor[HTML]{EFEFEF}\textbf{0.0483} & \cellcolor[HTML]{EFEFEF}\textbf{0.0231} \\ \bottomrule
\end{tabular}}
\end{table}

\begin{table}[t]
\caption{\textbf{Ablation of training losses.} Performance comparison with and without the edge-length and Jacobian losses, on top of the per-vertex deformation loss. Best results per metric are highlighted.}
\label{tab:ablate loss}
\centering
\resizebox{\linewidth}{!}{
\begin{tabular}{@{}lcccccc@{}}
\toprule
Loss & CD-3D $\downarrow$ & CD-4D $\downarrow$ & CD-M $\downarrow$ & CLIP $\uparrow$ & LPIPS $\downarrow$ & DreamSim $\downarrow$ \\ \midrule
Vertex only & 0.0588 & 0.0861 & 0.1556 & 0.9724 & 0.0525 & 0.0274 \\
+ Edge \& Jac. & \cellcolor[HTML]{EFEFEF}\textbf{0.0573} & \cellcolor[HTML]{EFEFEF}\textbf{0.0853} & \cellcolor[HTML]{EFEFEF}\textbf{0.1539} & \cellcolor[HTML]{EFEFEF}\textbf{0.9753} & \cellcolor[HTML]{EFEFEF}\textbf{0.0483} & \cellcolor[HTML]{EFEFEF}\textbf{0.0231} \\ \bottomrule
\end{tabular}
}
\end{table}

\subsection{Ablation Studies}

We evaluate the effectiveness of our key design choices. All ablations are conducted on 4-frame sequences sampled from ActionBench, with each variant trained for 100K iterations.

\paragraph{Encoder Input \& Modality.}
As described in Sec.~\ref{sec:encoder}, our encoder takes only the topology-aware representation of the anchor mesh and replicates it across all $T$ time steps to form the encoder input. Table~\ref{tab:ablate input} compares this design against an alternative that feeds a separately computed topology-aware representation for every frame. The per-frame variant yields no measurable accuracy gain, while inflating peak training memory by more than $50\%$. This validates the shared-anchor design as a favorable accuracy--efficiency trade-off.

We further ablate the two complementary modalities ingested by the encoder. As shown in Table~\ref{tab:ablate condition}, the 3D shape-latent prior contributes substantially to inter-frame correspondence, as removing it causes a pronounced drop across all metrics. Image features provide a smaller but consistent improvement, sharpening the encoder's semantic understanding of the sequence on top of the shape prior.

\paragraph{Loss Functions.} Table~\ref{tab:ablate loss} confirms the benefit of our edge-length and Jacobian losses, which yield consistent gains across all metrics on top of the per-vertex deformation loss. As illustrated in Fig.~\ref{fig:loss} both losses are defined over the anchor-mesh connectivity graph and therefore enjoy a larger effective receptive field than per-vertex supervision alone. They more effectively suppress geometric-collapse artifacts and preserve local structural fidelity.

\paragraph{Point Representation \& Global-Translation Prediction.} Since cases involving topology change or large global translation account for only a small fraction of the benchmark, we therefore turn to qualitative analysis. Fig.~\ref{fig:topo} shows that the topology-aware representation effectively prevents vertex entanglement during deformation, correctly disambiguating Euclidean-close yet geodesically distant regions. Fig.~\ref{fig:global} shows that the additional global-translation branch stabilizes the output under large translational motion and produces noticeably smoother surfaces.

\subsection{Additional Results}
\label{subsec: additional res}

\paragraph{Registration across Geometric Variations.} Fig.~\ref{fig:application}(a) demonstrates that \awesomemesh generalizes across geometric variations, such as registration across species. The registered outputs closely conform to the unregistered input shapes.
\paragraph{Motion Interpolation and Mesh Morphing.}
Benefiting from the global embedding-then-query paradigm (Sec.~\ref{sec:decoder}), the deformation decoder can be queried at arbitrary intermediate timestamps that the encoder has never observed. Fig.~\ref{fig:application}(b) shows motion-interpolation results between the anchor shape and multiple target motion states. The synthesized intermediate frames exhibit smooth temporal transitions while faithfully preserving object identity and local structure.

Fig.~\ref{fig:application}(c) further extends this capability to mesh morphing between two distinct objects, where our method produces smooth and coherent transitions from the anchor to the target. Together, these results demonstrate that our encoder learns continuous deformation trajectories from sparsely observed motion states, rather than merely reproducing the input frames.

\section{Conclusions}

In this work, we present \awesomemesh, a feed-forward non-rigid mesh registration framework that unifies heterogeneous mesh sequences into a topology-consistent sequence. Within a single model, \awesomemesh resolves the key bottlenecks of existing registration methods: the network runs at high speed, supports open-vocabulary objects, directly outputs per-vertex deformations, and processes variable-length sequences in a single pass. Extensive experiments show that our method achieves state-of-the-art performance across diverse object animations, and we further demonstrate its ability to generalize to motion interpolation and mesh morphing. We believe \awesomemesh will benefit the community and the broad range of downstream tasks that require persistent mesh identity over time.

\bibliographystyle{ACM-Reference-Format}
\bibliography{main}

\clearpage

\appendix

\section{Overview}

In this appendix, we first provide additional details on the network design in Appendix~\ref{app:net}, including an efficient implementation of the topology-aware representation, the time and 3D positional conditioning used in the encoder, and the positional encoding used in the decoder. We then describe additional processing details for the training and evaluation datasets in Appendix~\ref{app:data}. Next, Appendix~\ref{app:train} elaborates on the training objectives and introduces our training schemes. Appendix~\ref{app:metric} describes how the evaluation metrics are computed during evaluation. After that, Appendix~\ref{app: implementation details} provide the implementation details. To complement Table~\ref{tab:compare} in the main paper, Appendix~\ref{app:results} reports per-setting quantitative results. Finally, Appendix~\ref{app: limitation} discusses limitations and future work.

\section{Additional Network-Design Details}
\label{app:net}

\subsection{Topology-Aware Point Representation}

As shown in Eq.~\ref{eq:gcn} of the main paper, we enlarge the receptive field of each per-vertex feature by applying the adjacency operator $p$ times, yielding $\mathbf{A}^{p}$.

Although $\mathbf{A}$ itself is stored as a sparse matrix, naively materializing $\mathbf{A}^{p}$ via $p$ sparse--sparse multiplications (\ie repeatedly multiplying $\mathbf{A}$ by itself) produces an increasingly dense matrix and incurs memory cost quadratic in $N$ for large anchor meshes. We instead compute the propagated feature $\mathbf{A}^{p}\mathbf{H}^{(\ell)}$ through $p$ successive sparse--dense products (\ie multiplying the sparse matrix $\mathbf{A}$ with the dense matrix $\mathbf{H}^{(\ell)}$ at each step),
\begin{equation}
    \mathbf{A}^{p} \mathbf{H}^{(\ell)} \;=\; \mathbf{A}\big(\mathbf{A}(\cdots(\mathbf{A}\,\mathbf{H}^{(\ell)}))\big),
\end{equation}
which is mathematically equivalent yet keeps peak memory linear in the number of mesh edges.

\subsection{Motion Encoder}

\paragraph{Time conditioning.}
To make each token aware of the frame it belongs to, we inject the corresponding timestamp into the token features before every attention layer of every transformer block (the two cross-attention and two self-attention layers described in Sec.~\ref{sec:encoder} of the main paper). Specifically, we encode each timestamp $t$ with a sinusoidal embedding and route it through a FiLM~\cite{perez2018film} modulator. The modulator produces frame-specific scale and shift parameters $(\alpha_t, \beta_t)$ that modulate token activations before every attention layer.

\paragraph{Sparse-tensor structure and 3D positional encoding.}
Trellis.2~\cite{xiang2025native} stores its 3D shape latents in a sparse-tensor format, in which each token carries both a 3D coordinate and a feature vector. To remain natively compatible with this representation, we organize the encoder input in the same sparse-tensor format, where each point sampled from the topology-aware representation of the anchor mesh is stored together with its 3D coordinate, so that the encoder tokens themselves become sparse-tensor tokens. This allows the cross-attention with the Trellis.2 shape latents and the inter-/intra-frame self-attention to be carried out as native sparse attention.

To disambiguate sparse-tensor tokens that share similar features but reside at different 3D positions, we apply 3D rotary positional embedding (RoPE)~\cite{su2024roformer} during all sparse-attention computations. RoPE additionally injects relative spatial-distance information between tokens, helping the encoder reason about local geometry.

For the cross-attention with image features, we directly attend between the sparse-tensor tokens and the image features without any additional positional embedding.

\subsection{Deformation Decoder}

Since the global motion embedding $\mathbf{Z}$ produced by the encoder also consists of sparse-tensor tokens, a natural choice would be to reuse 3D RoPE when these tokens serve as keys and values for the decoder cross-attention (Eq.~\ref{eq:local_deform} of the main paper) with the queries in the local-deformation branch.

However, we empirically find that using explicit 3D Fourier positional embeddings~\cite{mildenhall2021nerf} of the associated 3D coordinates yields better performance. We attribute this to the fact that RoPE encodes only \emph{relative} spatial relations between tokens, whereas Fourier embeddings preserve \emph{absolute} position information, which is more informative for deformation reconstruction in which absolute coordinates matter. We therefore augment the key/value tokens of the local-deformation branch as follows:
\begin{equation}
\widetilde{\mathbf{Z}}
= \big[\,\mathbf{Z} \,\big\|\, \gamma(\mathbf{p}_{\mathbf{Z}} - \bar{\mathbf{v}}^{a})\big],
\end{equation}
where $\mathbf{p}_{\mathbf{Z}}$ denotes the 3D coordinate associated with each token in $\mathbf{Z}$, and $\gamma(\cdot)\!:\!\mathbb{R}^{3}\!\to\!\mathbb{R}^{D_{\gamma}}$ is a Fourier embedding that lifts a 3D point into a multi-frequency sinusoidal basis of dimension $D_{\gamma}$. Since our decoder decouples global and local transformations into two separate branches (Sec.~\ref{sec:decoder} of the main paper), we center all coordinates by the anchor centroid $\bar{\mathbf{v}}^{a}$ before applying $\gamma$, which removes the global-translation bias and makes the local-deformation branch invariant to $\Delta\mathbf{c}^{(t)}$.

\section{Additional Dataset Details}
\label{app:data}

\paragraph{Training dataset.}
We use the animation subset of Texverse~\cite{zhang2025texverse} as our training data. For each animated sequence, we extract a contiguous window of $L{=}32$ frames. The per-frame structured shape latent $\mathbf{S}_{t}$ is produced by the Trellis.2~\cite{xiang2025native} data-processing pipeline, which voxelizes each mesh into an O-Voxel representation~\cite{zhang2025texverse} and encodes it into a compact sparse latent grid (sparse-tensor format). The per-frame image-conditioning input $\mathbf{I}_{t}$ is an RGB rendering from a fixed front view. We apply four quality filters to ensure training stability: (1) meshes with more than $150$K vertices are discarded to bound memory consumption; (2) static sequences (whose vertices remain unchanged across the sampled frames) are removed, since they provide no deformation signal; (3) shape latents exceeding $30$K voxels are excluded; and (4) samples with missing files, corrupted meshes, or NaN vertex values are dropped. The remaining $\sim$$30$K sequences are used for training.

\paragraph{Evaluation dataset.}
We evaluate on ActionBench~\cite{sabathier2026actionmesh}. ActionBench contains $128$ animation sequences spanning a broad range of object categories. For each sequence, the benchmark provides per-frame guidance images and ground-truth point clouds whose points correspond consistently across frames. Since the benchmark does not directly provide mesh sequences, we use the image-to-mesh network of ActionMesh~\cite{sabathier2026actionmesh} to generate the input mesh sequence to be registered. For the experiments, we linearly sample frames at different temporal intervals to construct variable-input settings: 2 frames with a large time gap, 4 frames with a moderate time gap, and 8 frames with denser temporal coverage.

\section{Additional Training Details}
\label{app:train}

\subsection{Training Objectives}
\label{app:train_obj}

\paragraph{Global and local deformation loss.} 
We provide explicit definitions for the ground-truth supervision targets $\Delta\mathbf{c}^{*(t)}$ and $\Delta\mathbf{r}_{i}^{*(t)}$ used by the global- and local-deformation losses,
\begin{equation}
    \Delta\mathbf{c}^{*(t)} = \bar{\mathbf{v}}^{*(t)} - \bar{\mathbf{v}}^{a},
    \qquad
    \Delta\mathbf{r}_i^{*(t)} = \bigl(\mathbf{v}_i^{*(t)} - \bar{\mathbf{v}}^{*(t)}\bigr) - \bigl(\mathbf{v}_i^{a} - \bar{\mathbf{v}}^{a}\bigr),
\end{equation}
where $\mathbf{v}_i^{a}$ and $\mathbf{v}_i^{*(t)}$ are the positions of anchor vertex $i$ in the anchor frame and at the ground-truth target frame $t$, respectively, and $\bar{\mathbf{v}}^{a}$ and $\bar{\mathbf{v}}^{*(t)}$ are the corresponding full-mesh centroids. Subtracting both centroids supervises the local residual in a centroid-removed coordinate frame, so that it is fully decoupled from the global shift.

\paragraph{Edge and Jacobian losses.} The edge-length loss $\mathcal{L}_{\text{edge}}$ and the Jacobian loss $\mathcal{L}_{\text{jac}}$ defined in Sec.~\ref{sec:training} of the main paper rely on a connectivity graph built over a subset of anchor vertices. To balance supervision density and computational cost, we start from the decoder query points and run a multi-hop breadth-first search (BFS) along the anchor-mesh edges to collect topologically linked vertices, yielding an expanded set $\mathcal{V}'$ of $N' = N_{\mathrm{q}} + N_{\text{extra}}$ points. On this expanded set we build a $k$-nearest face-adjacent neighbor graph $\mathcal{G} = (\mathcal{V}', \mathcal{E})$ in which every edge corresponds to an actual anchor-mesh edge, so that no spurious long-range shortcuts are introduced. Both $\mathcal{L}_{\text{edge}}$ and $\mathcal{L}_{\text{jac}}$ are evaluated on this graph.

For each vertex $i$ in the expanded set, the per-vertex Jacobian $\mathbf{J}_{i}$ defined in Sec.~\ref{sec:training} is estimated by Tikhonov-regularized least squares over its $k$ face-adjacent neighbors,
\begin{equation}
    \mathbf{J}_i
    = \operatorname*{arg\,min}_{\mathbf{J}\in\mathbb{R}^{3\times 3}}
      \sum_{j \in \mathcal{N}(i)}
      \bigl\|
        (\widetilde{\mathbf{v}}_j^{(t)} - \widetilde{\mathbf{v}}_i^{(t)})
        - \mathbf{J}(\mathbf{v}_j^{a} - \mathbf{v}_i^{a})
      \bigr\|_2^2
      + \lambda_{\text{reg}}\|\mathbf{J}\|_F^2,
\end{equation}
where $\mathcal{N}(i)$ denotes the $k$ face-adjacent neighbors of vertex $i$ in $\mathcal{G}$, and $\lambda_{\text{reg}}$ is a small regularization weight that stabilizes the linear system in nearly degenerate neighborhoods. The ground-truth Jacobian $\mathbf{J}_{i}^{*}$ is obtained analogously from the ground-truth deformation.

\subsection{Training Schemes}

\begin{figure}[!t]
  \centering
   \includegraphics[width=\linewidth]{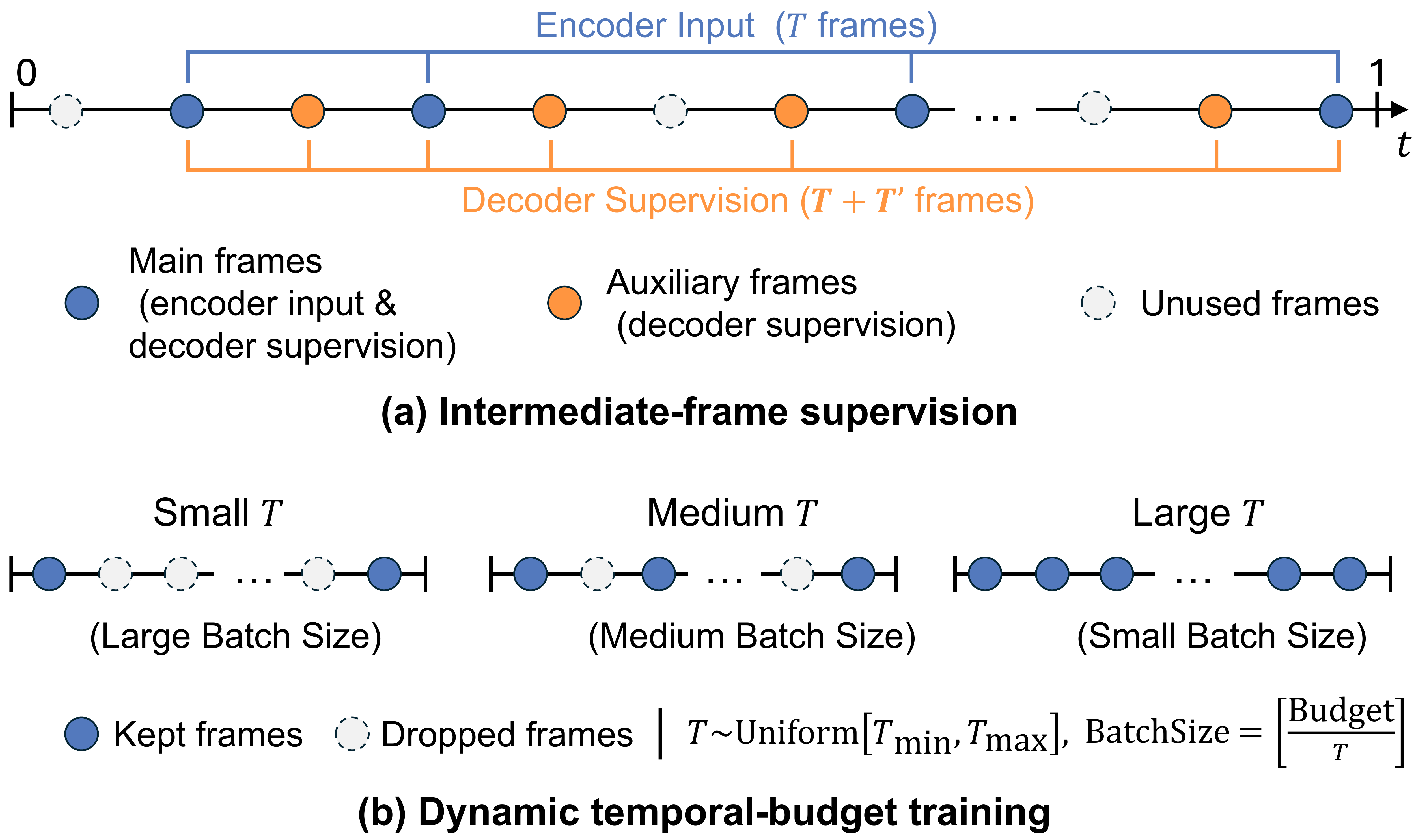}
   \caption{\textbf{Two of our training schemes.} Illustration of (i) intermediate-frame supervision, in which the decoder is supervised at both main and auxiliary timestamps while the encoder observes only the main subset, and (ii) dynamic temporal-budget training, in which the temporal length $T$ and the batch size $B$ are jointly resampled per iteration so that the per-GPU token count $\mathcal{B}=B\cdot T$ remains roughly constant.}
   \label{fig:train schemes}
\end{figure}

Beyond the architectural designs in the main paper, we adopt three training schemes that further enhance the model: an \emph{intermediate-frame supervision} scheme that equips the decoder with motion-interpolation ability at unseen timestamps; a \emph{dynamic temporal-budget} sampler that exposes the model to variable-length contexts so that the trained network generalizes across input lengths at inference; and a \emph{random image-dropout} scheme that randomly replaces the per-frame image features with an all-zero placeholder, allowing the network to gracefully handle inputs without accompanying images. The first two schemes are illustrated in Fig.~\ref{fig:train schemes}.

\paragraph{Intermediate-frame supervision.}
A subtle but important point in our training scheme is the distinction between the frames fed to the encoder and the timestamps at which the decoder is supervised. Each training sequence covers $L$ mesh frames whose frame indices are normalized to timestamps in $[0,1]$. From this pool, we draw two disjoint subsets at every iteration. A set of $T$ \emph{main} frames is passed to the motion encoder to construct $\mathbf{Z}$. An additional set of $T'$ \emph{auxiliary} frames is drawn from the remaining timestamps and is used \emph{only} for decoder supervision. The encoder, therefore, sees only a sparse subset of keyframes, yet the decoder is asked to reconstruct the full deformation at every main and auxiliary timestamp, including timestamps that the encoder has never observed. This explicitly trains the decoder to \emph{interpolate} the motion embedding across time, and allows the trained model to synthesize plausible mesh deformations at arbitrary intermediate timestamps at inference, supporting motion interpolation and mesh morphing.

\begin{table*}[t]
\renewcommand{\arraystretch}{1.5}
\caption{\textbf{Detailed performance comparison on ActionBench.} We report both geometric metrics (CD-3D, CD-4D, CD-Motion) and rendering-based visual metrics (CLIP, LPIPS, DreamSim) under each of the $2$-, $4$-, and $8$-frame input settings. We additionally report inference time, measured on $8$-frame inputs with a fixed $150$K vertices and $300$K faces per mesh frame. Best results are in \textbf{bold}, second-best are \underline{underlined}.}
\label{tab:full compare}
\centering
\resizebox{\linewidth}{!}{
\begin{tabular}{@{}l|ccc|ccc|ccc|ccc|ccc|ccc|c@{}}
\toprule
 & \multicolumn{3}{c|}{\textbf{CD-3D   $\downarrow$}} & \multicolumn{3}{c|}{\textbf{CD-4D   $\downarrow$}} & \multicolumn{3}{c|}{\textbf{CD-M $\downarrow$}} & \multicolumn{3}{c|}{\textbf{CLIP $\uparrow$}} & \multicolumn{3}{c|}{\textbf{LPIPS $\downarrow$}} & \multicolumn{3}{c|}{\textbf{DreamSim   $\downarrow$}} &  \\ \cmidrule(lr){2-19}
\multirow{-2}{*}{\textbf{Method}} & \textbf{2} & \textbf{4} & \textbf{8} & \textbf{2} & \textbf{4} & \textbf{8} & \textbf{2} & \textbf{4} & \textbf{8} & \textbf{2} & \textbf{4} & \textbf{8} & \textbf{2} & \textbf{4} & \textbf{8} & \textbf{2} & \textbf{4} & \textbf{8} & \multirow{-2}{*}{\textbf{\begin{tabular}[c]{@{}c@{}}Infer. \\ Time\end{tabular}}} \\ \midrule
TransferMatch~\cite{trappolini2021shape} & 0.0843 & 0.1000 & 0.1081 & 0.1410 & 0.1828 & 0.2047 & 0.6101 & 0.8939 & 1.0332 & 0.8283 & 0.7474 & 0.7072 & 0.1448 & 0.1715 & 0.1846 & 0.1785 & 0.2589 & 0.2991 & 14.2 \\
NDP~\cite{li2022non} & {\ul 0.0541} & 0.0546 & 0.0548 & 0.0799 & 0.0846 & 0.0885 & 0.1438 & 0.1761 & 0.1898 & 0.9736 & 0.9652 & 0.9610 & 0.0561 & 0.0638 & 0.0671 & 0.0265 & 0.0339 & 0.0370 & {\ul 5.9} \\
LNDP~\cite{li2022non} & 0.0610 & \cellcolor[HTML]{EFEFEF}\textbf{0.0487} & 0.0658 & 0.0873 & \cellcolor[HTML]{EFEFEF}\textbf{0.0646} & 0.1037 & 0.1450 & 0.1912 & 0.1917 & 0.9678 & 0.9584 & 0.9553 & 0.0983 & 0.1049 & 0.1078 & 0.0409 & 0.0500 & 0.0531 & 279.9 \\
DPF~\cite{prokudin2023dynamic} & \cellcolor[HTML]{EFEFEF}\textbf{0.0530} & {\ul 0.0531} & {\ul 0.0533} & \cellcolor[HTML]{EFEFEF}\textbf{0.0758} & {\ul 0.0832} & \cellcolor[HTML]{EFEFEF}\textbf{0.0870} & 0.1479 & 0.1831 & 0.2001 & 0.9785 & 0.9701 & 0.9659 & 0.0297 & \cellcolor[HTML]{EFEFEF}\textbf{0.0385} & \cellcolor[HTML]{EFEFEF}\textbf{0.0426} & 0.0157 & 0.0218 & 0.0248 & 70.2 \\
ClusterReg~\cite{zhao2024correspondence} & {\ul 0.0541} & 0.0546 & 0.0548 & 0.0781 & 0.0844 & 0.0880 & 0.1572 & 0.1978 & 0.2155 & 0.9085 & 0.8688 & 0.8484 & 0.0998 & 0.1156 & 0.1237 & 0.0788 & 0.1097 & 0.1250 & 42.6 \\
OAReg~\cite{zhao2025occlusion} & 0.0548 & 0.0565 & 0.0571 & 0.0784 & 0.0858 & 0.0899 & 0.1345 & 0.1649 & 0.1782 & 0.9748 & 0.9661 & 0.9617 & 0.0580 & 0.0655 & 0.0688 & 0.0263 & 0.0352 & 0.0387 & 25.3 \\
ActionMesh~\cite{sabathier2026actionmesh} & 0.0577 & 0.0570 & \cellcolor[HTML]{EFEFEF}\textbf{0.0532} & 0.0785 & 0.0869 & {\ul 0.0873} & 0.1325 & 0.1615 & 0.1731 & 0.9714 & 0.9669 & {\ul 0.9738} & 0.0958 & 0.0951 & 0.0553 & 0.0319 & 0.0345 & \cellcolor[HTML]{EFEFEF}\textbf{0.0196} & 7.7 \\ \midrule
Ours (w/o img) & 0.0566 & 0.0571 & 0.0577 & 0.0769 & 0.0853 & 0.0897 & {\ul 0.1269} & {\ul 0.1545} & {\ul 0.1666} & {\ul 0.9846} & {\ul 0.9776} & 0.9737 & {\ul 0.0267} & 0.0440 & 0.0518 & {\ul 0.0141} & {\ul 0.0208} & 0.0242 & \cellcolor[HTML]{EFEFEF} \\
Ours & 0.0562 & 0.0567 & 0.0572 & {\ul 0.0765} & 0.0850 & 0.0888 & \cellcolor[HTML]{EFEFEF}\textbf{0.1245} & \cellcolor[HTML]{EFEFEF}\textbf{0.1524} & \cellcolor[HTML]{EFEFEF}\textbf{0.1642} & \cellcolor[HTML]{EFEFEF}\textbf{0.9855} & \cellcolor[HTML]{EFEFEF}\textbf{0.9781} & \cellcolor[HTML]{EFEFEF}\textbf{0.9744} & \cellcolor[HTML]{EFEFEF}\textbf{0.0254} & {\ul 0.0435} & {\ul 0.0507} & \cellcolor[HTML]{EFEFEF}\textbf{0.0126} & \cellcolor[HTML]{EFEFEF}\textbf{0.0197} & {\ul 0.0230} & \multirow{-2}{*}{\cellcolor[HTML]{EFEFEF}\textbf{3.1}} \\ \bottomrule
\end{tabular}
}
\end{table*}

\paragraph{Dynamic temporal-budget training.}
Fixing the input length $T$ during training would induce a train--test distribution mismatch, since at inference, the user may supply sequences of arbitrary length. We therefore adopt a \emph{dynamic-budget} strategy that randomizes $T$ per iteration while keeping the GPU memory footprint roughly constant. We fix a per-GPU token budget $\mathcal{B} = B \cdot T$ (batch size $\times$ temporal frames) and, at each step, sample
\begin{equation}
    T \sim \mathrm{Uniform}[T_{\min},\, T_{\max}],
    \qquad
    B = \lfloor \mathcal{B} / T \rfloor,
\end{equation}
so that a smaller $T$ yields a proportionally larger effective batch size, and vice versa. When $T < T_{\max}$, we temporally subsample the main frames while always retaining the first and last frames to preserve the full temporal span. The random seed governing $T$ is synchronized across all distributed processes to ensure identical tensor shapes across ranks. Combined with the intermediate-frame supervision above, this strategy exposes the model to temporal contexts ranging from two-frame pairs to long sequences, enabling the network to generalize to variable-length inputs at inference without any architectural change.

\paragraph{Random image dropout.}
In real-world deployments, the input mesh sequence may not always be accompanied by RGB images (although such images can be readily obtained from off-the-shelf image-generation models). To make our network robust to this scenario, we train it to handle both the with-image and without-image input modes. Inspired by classifier-free guidance~\cite{ho2022classifier} in image-generation models, at each training iteration, we replace the per-frame image features $\mathbf{I}_{t}$ with an all-zero placeholder with probability $0.5$. This allows the network to fall back gracefully to a vision-free mode when reference images are unavailable at inference.

\section{Evaluation Metric Details}
\label{app:metric}

For the visual metrics (LPIPS, CLIP Similarity, and DreamSim), we render normal maps for each mesh frame from four azimuth angles ($0^\circ$, $90^\circ$, $180^\circ$, and $270^\circ$). We first average the metric values across these four views, and then average over all frames in the sequence to obtain the final score.

For the geometric metrics (CD-3D, CD-4D, and CD-Motion) introduced in Sec.~\ref{sec:experiments} of the main paper, we first sample dense point clouds from the predicted mesh sequences and align them with the ground-truth point clouds using Iterative Closest Point (ICP). ICP is performed on $10$K-point subsamples, while the symmetric Chamfer distance is computed on denser $100$K-point samples.

\paragraph{CD-3D.} We estimate an independent per-frame ICP alignment between the predicted and ground-truth point clouds at every frame. CD-3D is then the Chamfer distance between the per-frame-aligned prediction and the ground truth, averaged across all frames. As the alignment is recomputed at every frame, CD-3D factors out global pose drift and isolates per-frame shape accuracy.

\paragraph{CD-4D.} We estimate a single ICP alignment from the first frame and apply it uniformly to every subsequent predicted frame. CD-4D is the Chamfer distance after this unified alignment, averaged across all frames. Holding the alignment fixed over time penalizes both shape error and accumulated temporal pose drift, jointly reflecting shape fidelity and temporal pose consistency.

\paragraph{CD-Motion.} CD-Motion further probes motion coherence by \emph{synchronizing} the surface sampling across frames. We compute face indices and barycentric weights once on the first predicted frame and reuse them on every subsequent frame, so that each sampled point traces a temporal trajectory on the surface. After applying the unified ICP transform, we establish bidirectional nearest-neighbor correspondences between the predicted and ground-truth point clouds at the first frame, propagate these correspondences across the entire sequence, and report the symmetric mean per-trajectory Euclidean distance. This penalizes temporally inconsistent surface motion even when per-frame shapes are individually accurate.

\section{Implementation details}
\label{app: implementation details}
The topology-aware point representation uses $L{=}4$ residual GCN-style layers, each aggregating neighbor information through a $p{=}4$-hop adjacency, which together provide a sufficiently large receptive field over the anchor mesh. For the encoder input, we FPS-sample $N_{\mathrm{ds}}{=}2{,}048$ points from the anchor-mesh representation $\mathbf{H}_{a}$. We adopt the recent Trellis.2~\cite{xiang2025native} 3D foundation model to obtain the shape-latent prior of each mesh frame, and DINOv3~\cite{simeoni2025dinov3} to extract image features. The encoder consists of $5$ transformer blocks. To bound memory usage when processing meshes with very large vertex counts, at each training iteration we sample $N_{\mathrm{q}}{=}4{,}096$ anchor vertices per mesh frame (half via FPS and half via uniform random sampling) as decoder query points for vertex-level supervision. For the geometry graph used by the edge-length and Jacobian losses (Appendix~\ref{app:train}), we expand this query set via a $2$-hop BFS along anchor-mesh edges (adding up to $N_{\text{extra}}{=}4{,}096$ extra vertices) and build a $k{=}8$ face-adjacent nearest-neighbor graph on the expanded set. The dynamic-budget sampler uses $\mathcal{B}{=}16$ with $T_{\min}{=}2$ and $T_{\max}{=}16$ encoder frames per iteration, yielding effective per-GPU batch sizes between $1$ and $8$. For intermediate-frame supervision, we draw the same number of auxiliary frames from the remaining timestamps to provide additional decoder supervision.

We optimize with AdamW~\cite{loshchilov2018decoupled} at a learning rate of $4{\times}10^{-4}$. The loss weights are set to $\lambda_{\text{global}}{=}\lambda_{\text{local}}{=}1$, $\lambda_{\text{edge}}{=}0.5$, and $\lambda_{\text{jac}}{=}0.05$. Training is conducted on $32$ NVIDIA A100 GPUs and converges after $200\text{K}$ iterations.

\section{Additional Quantitative Results}
\label{app:results}

Table~\ref{tab:full compare} complements Table~\ref{tab:compare} of the main paper by providing per-setting quantitative results of each method under the $2$-, $4$-, and $8$-frame input settings.

\section{Limitations and Future Work.}
\label{app: limitation}

Despite its strong performance across diverse scenarios, \awesomemesh still has several limitations. First, although the network internalizes topology information, the final registration fidelity remains dependent on the quality of the anchor mesh. If the anchor fails to faithfully capture the connectivity of the underlying object structure, the model cannot, by itself, recover the missing connectivity during registration. A promising remedy is to first reconstruct an A-pose anchor mesh in which the connectivity is clearly expressed. Second, the network struggles under extreme motion intensities. When two consecutive frames differ by a large rigid transformation, \eg, a near-symmetric human character rotating $180^\circ$, it becomes difficult to disambiguate correspondence, and the encoder loses track of the underlying motion. Incorporating stronger cues, such as multi-view renderings, is a promising direction for addressing this issue. Finally, since our current model is trained on object-centric 4D animation data, scaling to scene-level registration with multiple interacting objects remains an open direction, where the main bottleneck is likely to be data scarcity.

\end{document}